\documentclass[preprint,12pt]{elsarticle}

\biboptions{sort&compress}
\usepackage{graphicx}      
\usepackage{amssymb}
\usepackage{amsthm}

\usepackage{amsmath}
\usepackage{epstopdf}
\usepackage{subfigure}
\usepackage{comment}
\usepackage{multirow}
\usepackage{ctable}
\usepackage[flushleft]{threeparttable}
\usepackage{array}
\journal{Neural Networks}

\begin{document}

\begin{frontmatter}



\title{Edge Detectors Can Make Deep Convolutional Neural Networks More Robust}


\author[label1,label2]{Jin Ding\corref{cor1}}
\ead{jding@zust.edu.cn}
\author[label1]{Jie-Chao Zhao}
\ead{zjiechao@126.com}
\author[label1]{Yong-Zhi Sun}
\ead{sunyongzhi@hotmail.com}
\author[label1]{Ping Tan\corref{cor1}}
\ead{115011@zust.edu.cn}
\author[label1]{Jia-Wei Wang}
\ead{13175040964@163.com}
\author[label2,label3]{Ji-En Ma}
\ead{majien@zju.edu.cn}
\author[label2,label3]{You-Tong Fang}
\ead{youtong@zju.edu.cn}

\address[label1]{School of Automation and Electrical Engineering \& Key Institute of Robotics of Zhejiang Province, Zhejiang University of Science and Technology, Hangzhou, 310023, China}
\address[label2]{State Key Laboratory of Fluid Power and Mechatronic Systems, Zhejiang University, Hangzhou 310027, China}
\address[label3]{School of Electrical Engineering, Zhejiang University, Hangzhou, 310027, China}

\cortext[cor1]{\textbf{Corresponding author}}

\begin{abstract}
Deep convolutional neural networks (DCNN for short) are vulnerable to examples with small perturbations. Improving DCNN's robustness is of great significance to the safety-critical applications, such as autonomous driving and industry automation. Inspired by the principal way that human eyes recognize objects, i.e., largely relying on the shape features, this paper first employs the edge detectors as layer kernels and designs
a binary edge feature branch (BEFB for short) to learn the binary edge features, which can be easily integrated into any popular backbone. The four edge detectors can learn the horizontal, vertical, positive diagonal, and negative diagonal edge features, respectively, and the branch is stacked by multiple Sobel layers (using edge detectors as kernels) and one threshold layer. The binary edge features learned by the branch, concatenated with the texture features learned by the backbone, are fed into the fully connected layers for classification. We integrate the proposed branch into VGG16 and ResNet34, respectively, and conduct experiments on multiple datasets. Experimental results demonstrate the BEFB is lightweight and has no side effects on training. And the accuracy of the BEFB integrated models is better than the original ones on all datasets when facing FGSM, PGD, and C\&W attacks. Besides, BEFB integrated models equipped with the robustness enhancing techniques can achieve better classification accuracy compared to the original models. The work in this paper for the first time shows it is feasible to enhance the robustness of DCNNs through combining both shape-like features and texture features.

\end{abstract}

\begin{keyword}
learnable edge detectors \sep binary edge feature branch \sep Sobel layer \sep threshold layer \sep adversarial robustness \sep deep convolutional neural networks
\end{keyword}

\end{frontmatter}
%
%


\section{Introduction}
\begin{figure}[h!]
\centering
\subfigure[Clean examples]{\includegraphics[width=0.3\textwidth]{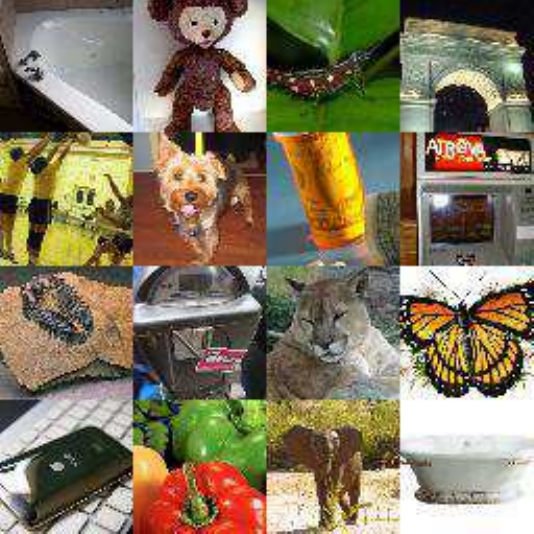}\label{fig:ORI}}
\hspace{0.05\textwidth}
\subfigure[Adversarial examples]{\includegraphics[width=0.3\textwidth]{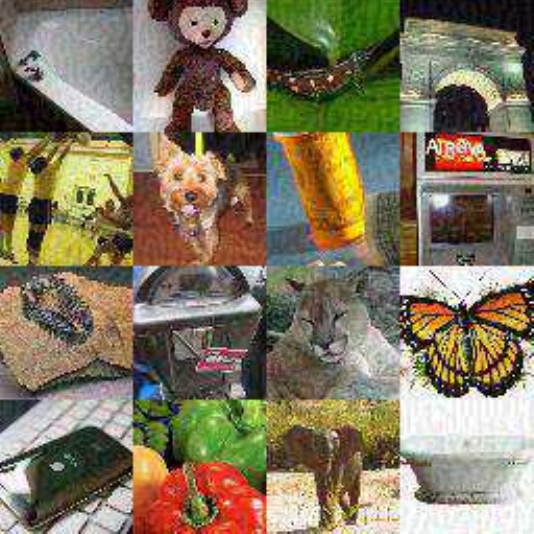}\label{fig:AE}}
\caption{\textbf{Clean examples and adversarial examples}.}
\label{fig:ORIAE}
\end{figure}

It is well known that deep convolutional neural networks (DCNN for short) can be fooled by examples with small perturbations \cite{Goodfellow2015ICLR, Szegedy2014ICLR, Tao2021IJCAI, Zhang2019TNNLS, Serban2020CSUR}, which brings the potential hazards when applying DCNNs to the safety-critical applications, e.g., autonomous driving, airport security, and industry automation. Therefore, it is of great significance to improve the robustness of DCNNs. Fig.~\ref{fig:ORIAE} shows several clean examples and their adversarial counterparts. Adversarial examples (AE for short) in Fig.~\ref{fig:AE} are with small noises which can induce DCNNs to make wrong decisions. However, these noises can be filtered by human eyes easily. Researches show that the principal way human beings recognize the objects is largely relying on the shape features \cite{Davitt2014JEP, Sigurdardottir2014JEP}, and that's why human beings are not susceptible to the noise imposed in Fig.~\ref{fig:AE}. Inspired by this, it is natural to ask, is it possible to make the DCNNs more robust by learning the shape-like features?

\begin{figure}[h!]
\centering
\subfigure[Thresholded edge images for clean examples]{\includegraphics[width=0.3\textwidth]{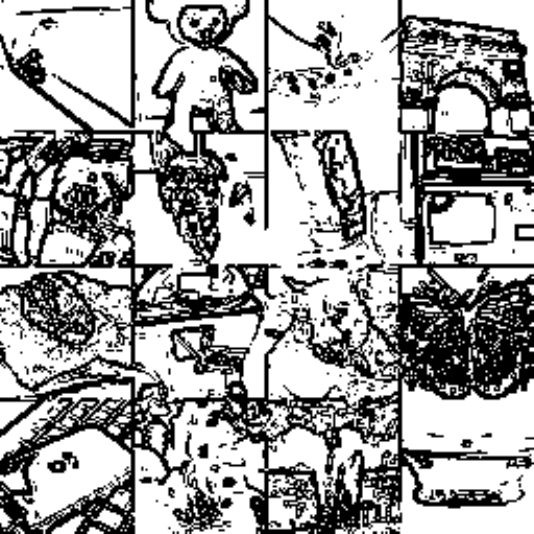}\label{fig:ORIBI}}
\hspace{0.05\textwidth}
\subfigure[Thresholded edge images for adversarial examples]{\includegraphics[width=0.3\textwidth]{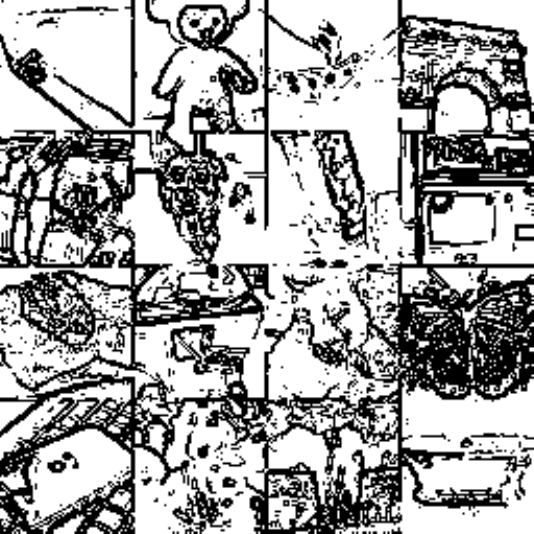}\label{fig:AEBI}}
\caption{\textbf{Thresholded edge images of Fig.~\ref{fig:ORIAE}}.}
\label{fig:ORIAEBI}
\end{figure}

Fig.~\ref{fig:ORIAEBI} shows the thresholded edge images of Fig.~\ref{fig:ORIAE}. From the figure, it can be observed that the thresholded edge images for Fig.~\ref{fig:ORI} and Fig.~\ref{fig:AE} are almost the same. It prompts us that, enhancing the robustness of DCNNs can be achieved by using the binary edge features which can be seen as a kind of shape features. To this end, in this paper, four learnable edge detectors are designed and taken as layer kernels, which can be used to extract horizontal, vertical, positive diagonal, and negative diagonal edge features, respectively \cite{Kittler1983IVC, Xiao2022JOS, Yasir2022Science}. And based on the edge detectors, a binary edge feature branch (BEFB for short) is proposed, which is staked by multiple Sobel layers and one threshold layer. The Sobel layers employ the edge detectors as kernels, and the threshold layer turns the output of the last Sobel layer to the binary features. The binary features concatenated with the texture features learnt by any popular backbone are then fed into fully connected layers for classification. To deal with the zero gradient of threshold layer, which makes the weights in BEFB unable to update using the chain rule, STE technique \cite{Bengio2013Arxiv, Yang2022ICML, Vanderschueren2023WACV} is employed. In addition, to take obfuscated gradient effect raised by \cite{Athalye2018ICML, Yue2023USENIX, Huang2022MSML} into consideration, zero gradient or one center-translated sigmoid activation function are also used for generating AEs. In the experiments, we integrate BEFB into VGG16 \cite{Simonyan2014ICLR} and ResNet34 \cite{He2016CVPR, He2020TNNLS}, respectively, and conduct experiments on multiple datasets. The results demonstrate BEFB has no side effects on model training, and BEFB integrated models can achieve better accuracy under FGSM \cite{Goodfellow2015ICLR}, PGD \cite{Kurakin2017ICLR}, and C\&W \cite{Carlini2017SP} attacks compared to the original ones. Furthermore, we combine the BEFB integrated models with two popular robustness enhancing techniques-\--AT(abbreviation for adversarial training) \cite{Madry2018ICLR} and PCL(abbreviation for prototype conformity loss) \cite{Mustafa2021TPAMI}, respectively, and find BEFB integrated models can achieve better accuracy than the original models as well.

The contributions of this paper can be summarized as follows:
\begin{enumerate}
  \item Inspired by the principal way that the human beings recognize objects, we design four learnable edge detectors and propose BEFB to make DCNNs more robust.
  \item BEFB can be easily integrated into any popular backbone, and has no side effects on model training. Extensive experiments on multiple datasets show BEFB integrated models can achieve better accuracy under FGSM, PGD, and C\&W attacks when compared to the original models.
  \item For the first time, we show it is feasible to make DCNNs more robust by combining the shape-like features and texture features.
\end{enumerate}

The organization of the paper is as follows. Section~\ref{sec:RW} briefly reviews the related works, and Section~\ref{sec:BEFB} describes the details of BEFB. In Section~\ref{sec:Expe} and \ref{sec:Dis}, the experiments on multiple datasets are conducted, and the discussions are made. Finally, in Section~\ref{sec:Con}, the concluding remarks are given.


\section{Related Work}\label{sec:RW}
\textbf{Robustness Enhancing with AT.}
AT improves the robustness of the DCNNs by generating AEs as training samples during optimization.
Tsipras \emph{et al.} \cite{Tsipras2019ICLR} found there exists a tradeoff between robustness and standard accuracy of the models generated by AT, due to the robust classifiers learning a different feature representation compared to the clean classifiers.
Madry \emph{et al.} \cite{Madry2018ICLR} proposed an approximated solution framework to the optimization problems of AT, and found PGD-based AT can produce models defending themselves against the first-order adversaries.
Kannan \emph{et al.} \cite{Kannan2018NIPS} investigated the effectiveness of AT at the scale of ImageNet \cite{Deng2009CVPR}, and proposed a logit pairing AT training method to tackle the tradeoff between robust accuracy and clean accuracy.
Wong \emph{et al.} \cite{Wong2020ICLR} accelerated the training process using FGSM attack with random initialization instead of PGD attack \cite{Kurakin2017ICLR}, and reached significantly lower cost.
Xu \emph{et al.} \cite{Xu2022NIPS} proposed a novel attack method which can make a stronger perturbation to the input images, resulting in the robustness of models by AT using this attack method is improved.
Li \emph{et al.} \cite{Li2022CVPR} revealed a link between fast growing gradient of examples and catastrophic overfitting during robust training, and proposed a subspace AT method to mitigate the overfitting and increase the robustness.
Dabouei \emph{et al.} \cite{Dabouei2022ECCV} found the gradient norm in AT is higher than natural training, which hinders the training convergence of outer optimization of AT. And they proposed a gradient regularization method to improve the performance of AT.

\textbf{Robustness Enhancing without AT.}
Non-AT robustness enhancing techniques can be categorized into part-based models \cite{Li2023TPAMI, Sitawarin2023ICLR}, feature vector clustering \cite{Mustafa2021TPAMI, Seo2023CVIU}, adversarial margin maximization \cite{Yan2019TPAMI, Guo2022TPAMI}, etc.
Li \emph{et al.} \cite{Li2023TPAMI} argued one reason that DCNNs are prone to be attacked is they are trained only on category labels, not on the part-based knowledge as humans do. They proposed an object recognition model, which first segments the parts of objects, scores the segmentations based on human knowledge, and final outputs the classification results based on the scores.
This part-based model shows better robustness than classic recognition models across various attack settings.
Sitawarin \emph{et al.} \cite{Sitawarin2023ICLR} also thought richer annotation information can help learn more robust features. They proposed a part segmentation model with a head classifier trained end-to-end. The model first segments objects into parts, and then makes predictions based on the parts.
Mustafa \emph{et al.} \cite{Mustafa2021TPAMI} stated that making feature vectors in the same class closer and centroids of different classes more separable can enhance the robustness of DCNNs. They added PCL to the conventional loss function, and designed an auxiliary function to reduce the dimensions of convolutional feature maps.
Seo \emph{et al.} \cite{Seo2023CVIU} proposed a training methodology that enhance the robustness of DCNNs through a constraint that applies a class-specific differentiation to the feature space. The training methodology results in feature representations with a small intra-class variance and large inter-class variances, and can improve the adversarial robustness notably.
Yan \emph{et al.} \cite{Yan2019TPAMI} proposed an adversarial margin maximization method to improve DCNNs' generalization ability. They employed the deepfool attack method \cite{Moosavi2016CVPR} to compute the distance between an image sample and its decision boundary. This learning-based regularization can enhance the robustness of DCNNs as well.

Note that, BEFB integrated models proposed in this paper can be easily combined with above-mentioned robustness enhancing techniques, such as AT \cite{Madry2018ICLR} and PCL \cite{Mustafa2021TPAMI}, and can achieve better classification accuracy than the original models.

\section{Proposed Approach}\label{sec:BEFB}
In this section, we first introduce four learnable edge detectors, and then illustrate a binary edge feature branch (BEFB for short).

\subsection{Edge Detectors}
\begin{figure}[h!]
\centering
\subfigure[Horizontal edge detector]{\includegraphics[width=0.3\textwidth]{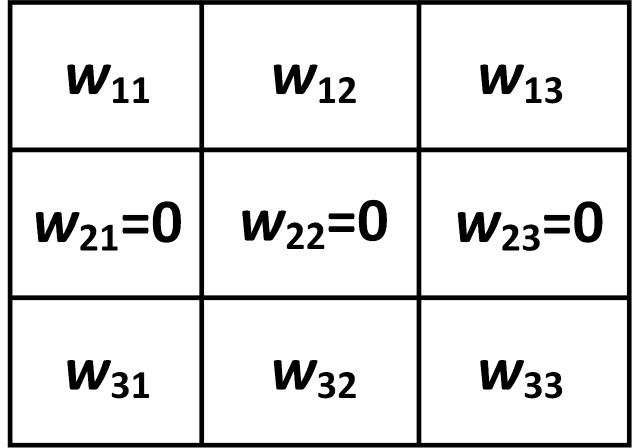}\label{fig:Khori}}
\hspace{0.1\textwidth}
\subfigure[Vertical edge detector]{\includegraphics[width=0.3\textwidth]{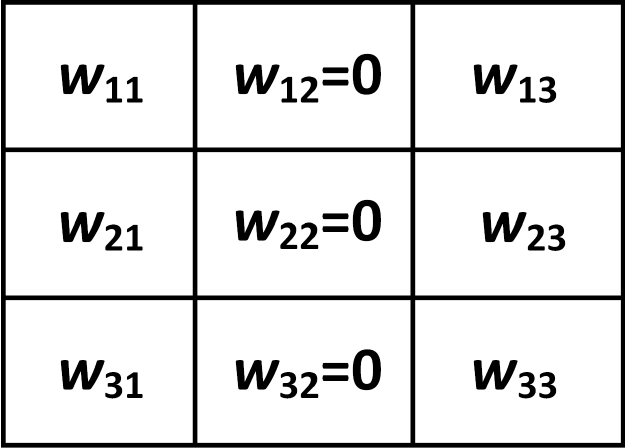}\label{fig:Kvert}}
\subfigure[Positive diagonal edge detector]{\includegraphics[width=0.3\textwidth]{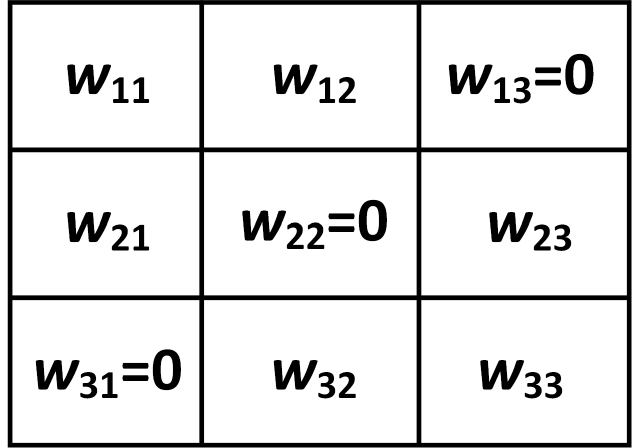}\label{fig:Kpodi}}
\hspace{0.1\textwidth}
\subfigure[Negative diagonal edge detector]{\includegraphics[width=0.3\textwidth]{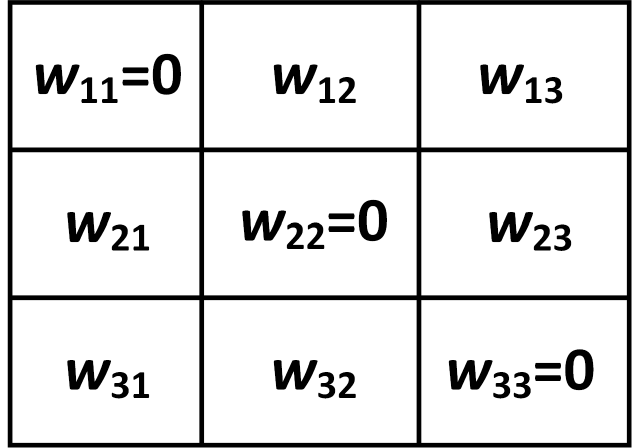}\label{fig:Knedi}}
\caption{\textbf{Four types of edge detectors}.}
\label{fig:Kernel}
\end{figure}

Inspired by the fact that shape features is the main factor relied on by human beings to recognize objects, here, we design four learnable edge detectors which can extract horizontal edge features, vertical edge features, positive diagonal edge features, and negative diagonal edge features, respectively. These four learnable edge detectors can be taken as layer kernels and are shown in Fig.~\ref{fig:Kernel}.

For horizontal edge detector in Fig.~\ref{fig:Khori}, we have
\begin{equation}
\label{eq:Khori}
w_{ij}\left\{
\begin{array}{ll}
\in \textrm{[0, 1]} & \mbox{if \emph{i} = 1, \emph{j} = 1, 2, 3},\\
= \textrm{0} & \mbox{if \emph{i} = 2, \emph{j} = 1, 2, 3},\\
\in \textrm{[-1, 0]} & \mbox{if \emph{i} = 3, \emph{j} = 1, 2, 3},\\
\end{array}
\right.
\end{equation}

For vertical edge detector in Fig.~\ref{fig:Kvert}, we have
\begin{equation}
\label{eq:Kvert}
w_{ij}\left\{
\begin{array}{ll}
\in \textrm{[0, 1]} & \mbox{if \emph{j} = 1, \emph{i} = 1, 2, 3},\\
= \textrm{0} & \mbox{if \emph{j} = 2, \emph{i} = 1, 2, 3},\\
\in \textrm{[-1, 0]} & \mbox{if \emph{j} = 3, \emph{i} = 1, 2, 3},\\
\end{array}
\right.
\end{equation}

For positive diagonal edge detector in Fig.~\ref{fig:Kpodi}, we have
\begin{equation}
\label{eq:Kpodi}
w_{ij}\left\{
\begin{array}{ll}
\in \textrm{[0, 1]} & \mbox{if (\emph{i}, \emph{j}) $\in$ \{(1, 1), (1, 2), (2, 1) \}},\\
= \textrm{0} & \mbox{if (\emph{i}, \emph{j}) $\in$ \{(1, 3), (2, 2), (3, 1) \}},\\
\in \textrm{[-1, 0]} & \mbox{if (\emph{i}, \emph{j}) $\in$ \{(2, 3), (3, 2), (3, 3) \}},\\
\end{array}
\right.
\end{equation}

For negative diagonal edge detector in Fig.~\ref{fig:Knedi}, we have
\begin{equation}
\label{eq:Knedi}
w_{ij}\left\{
\begin{array}{ll}
\in \textrm{[0, 1]} & \mbox{if (\emph{i}, \emph{j}) $\in$ \{(1, 2), (1, 3), (2, 3) \}},\\
= \textrm{0} & \mbox{if (\emph{i}, \emph{j}) $\in$ \{(1, 1), (2, 2), (3, 3) \}},\\
\in \textrm{[-1, 0]} & \mbox{if (\emph{i}, \emph{j}) $\in$ \{(2, 1), (3, 1), (3, 2) \}},\\
\end{array}
\right.
\end{equation}

\subsection{Binary Edge Feature Branch}
\begin{figure}[h!]
    \centering
    \includegraphics[width=0.9\textwidth]{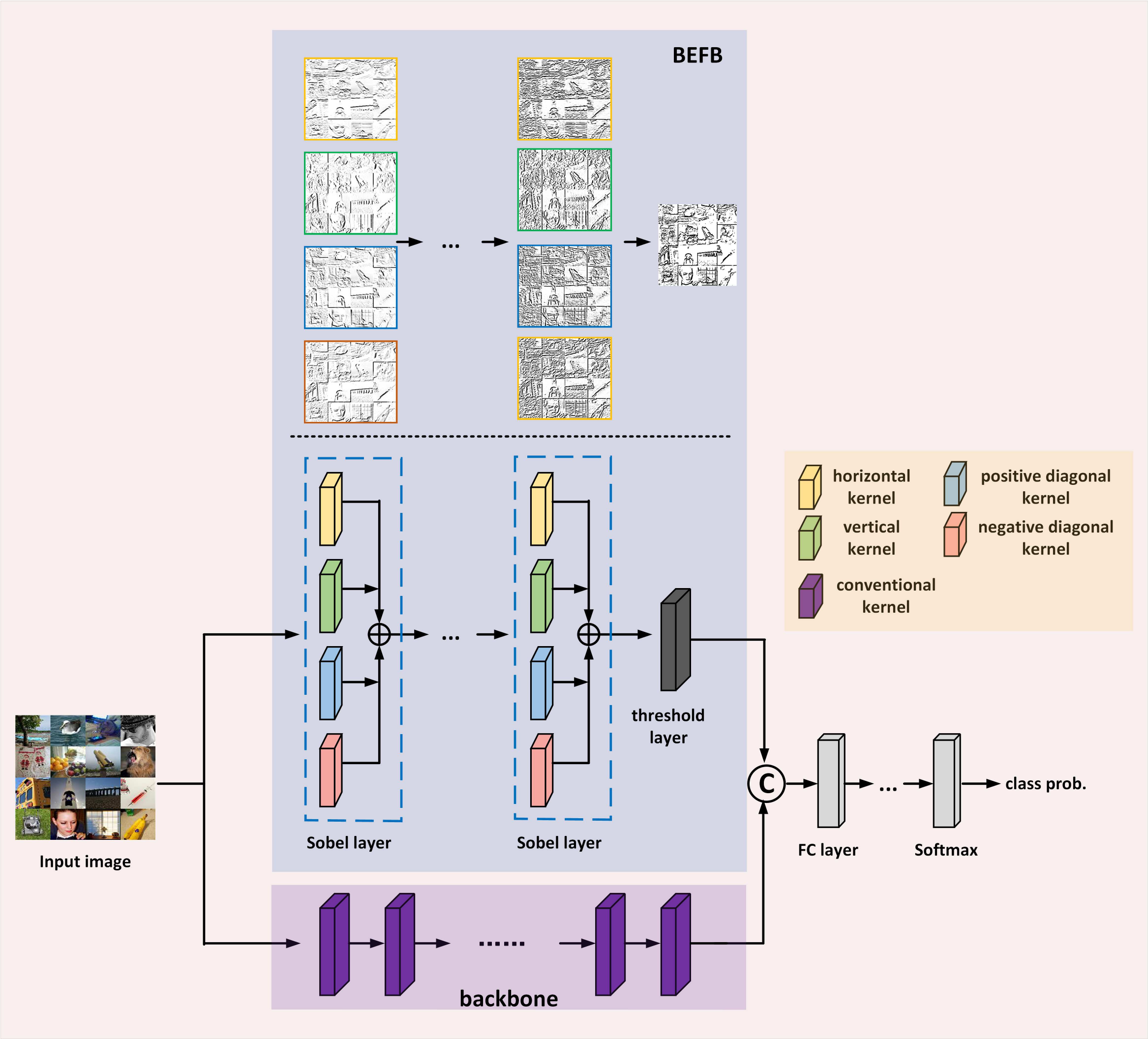}
    \caption{\textbf{The architecture of a BEFB integrated DCNN model.}}
    \label{fig:Archi}
\end{figure}

Based on the four learnable edge detectors, BEFB is proposed to extract the binary edge features of the images. BEFB is stacked by multiple Sobel layers and one threshold layer, and can be integrated into any popular backbone. The architecture of a BEFB integrated DCNN model is shown in Fig.~\ref{fig:Archi}.

\textbf{Sobel Layer. }
A Sobel layer is one single convolutional layer or multiple parallel convolutional layers using addition for fusion. In our experiments, for Sobel layer with one single convolutional layer, horizontal edge detector is used as its kernel. For Sobel layer with multiple parallel convolutional layers, we set the number of convolutional layers to four, and take horizontal, vertical, positive diagonal, and negative diagonal edge detectors as their kernels, respectively. Fig.~\ref{fig:Archi} shows a BEFB integrated model in which Sobel layer is with four parallel convolutional layers.

\textbf{Threshold Layer. }
The threshold layer in its nature is an activation layer. The activation function can be written as follows.


\begin{equation}
\label{eq:featureoutput}
x_{ikj}^{out}=\left\{
\begin{array}{ll}
1 & \mbox{if $x_{ikj}^{in}$ $\geq$ $t$ * $\max$($\textbf{x}_{i}^{in}$)},\\
0 & \mbox{if $x_{ikj}^{in}$ $<$ $t$ * $\max$($\textbf{x}_{i}^{in}$)},\\
\end{array}\\
\right.
\end{equation}

In Eq.~(\ref{eq:featureoutput}), $x_{ikj}^{in}$ is an element of the tensor $\textbf{x}^{in}$ $\in$ $R^{N\times P\times Q}$, which represents the feature map obtained by the last Sobel layer. $x_{ikj}^{out}$ is an element of the tensor $\textbf{x}^{out}$ $\in$ $R^{N\times P\times Q}$, which represents the binary feature map output by the threshold layer. $N$ is the number of channels of the feature map, and $P$ and $Q$ are the width and height of a channel. $i$ = 1, 2, ..., $N$, $k$ = 1, 2, ..., $P$, and $j$ = 1, 2, ..., $Q$. $t$ is a proportional coefficient and belongs to [0, 1]. $\max$($\textbf{x}_{i}^{in}$) represents the maximum value of channel $i$.
It is obvious to see that, the higher $t$, the less binary features obtained; the smaller $t$, the more binary features obtained.

The threshold layer turns the output of the last Sobel layer to the binary edge features, which concatenated with the texture features learnt by the backbone are fed into fully connected layers for classification. Note that, the gradient of activation function of Eq.~(\ref{eq:featureoutput}) is zero. To update the weights in BEFB, STE technique \cite{Bengio2013Arxiv, Yang2022ICML, Vanderschueren2023WACV} is employed.

\section{Experiments}\label{sec:Expe}
In this section, we integrate BEFB into VGG16 \cite{Simonyan2014ICLR} and ResNet34 \cite{He2016CVPR, He2020TNNLS}, respectively, and conduct the experiments on CIFAR-10 \cite{Krizhevsky2009}, MNIST \cite{Deng2012ISPM}, SVHN \cite{Yuval2011NIPS}, and TinyImageNet (TinyIN for short) \cite{Le2015Tiny} datasets.
We examine the effects of BEFB on model training, analyse obfuscated gradient effect \cite{Athalye2018ICML, Yue2023USENIX, Huang2022MSML} by using different activation functions of the threshold layer to generate AEs, and compare the accuracy of BEFB integrated models when facing FGSM \cite{Goodfellow2015ICLR}, PGD \cite{Kurakin2017ICLR}, and C\&W \cite{Carlini2017SP} attacks with original ones. Furthermore, we combine the BEFB integrated models with two robustness enhancing techniques-\--AT \cite{Madry2018ICLR} and PCL \cite{Mustafa2021TPAMI} to evaluate the classification accuracy.
All experiments are coded in Tensorflow with one TITAN XP GPU.

\subsection{Experimental Settings}
In the experiments, we adopt two types of BEFB integrated models. One is Sobel layer with a single convolutional layer, denoted as BEFB-single. The other is Sobel layer with four parallel convolutional layers, denoted as BEFB-multiple.
The settings of the number of Sobel layers $l$ and proportional coefficient $t$ of threshold layer are shown in table~\ref{tab:ltsettings}.
The settings of the $\epsilon$, $steps$, and $stepsize$ of FGSM and PGD are shown in table~\ref{tab:pgdsettings}.
Each model is run for five times, and the best value is recorded.

\begin{table}[ht]
\scriptsize
\begin{center}
\begin{threeparttable}

\begin{tabular}{l|cc|cc|cc|cc}

\hline
\multirow{2}*{} & \multicolumn{2}{c|}{\textbf{CIFAR-10}} & \multicolumn{2}{c|}{\textbf{MNIST}} & \multicolumn{2}{c|}{\textbf{SVHN}} & \multicolumn{2}{c}{\textbf{TinyIN}} \\
		\cline{2-9}
		~ & \textbf{$l$} & \textbf{$t$} & \textbf{$l$} & \textbf{$t$} & \textbf{$l$} & \textbf{$t$} & \textbf{$l$} & \textbf{$t$} \\
\hline
\textbf{VGG16-BEFB} & 2 & 0.8 & 2 & 0.8 & 2 & 0.8 & 3 & 0.6 \\
\textbf{ResNet34-BEFB} & 2 & 0.6 & 2 & 0.6 & 2 & 0.6 & 3 & 0.6 \\
\hline
\end{tabular}
\end{threeparttable}
\end{center}
\caption{The settings of the number of Sobel layers and proportional coefficient of Threshold layer.}
\label{tab:ltsettings}
\end{table}

\begin{table}[ht]
\scriptsize
\begin{center}
\begin{threeparttable}

\begin{tabular}{l|c|c|c|c}
\hline
{} & \textbf{CIFAR-10} & \textbf{MNIST} & \textbf{SVHN} & \textbf{TinyIN} \\
\hline
\textbf{$\epsilon$} & 8 & 80 & 8 & 8 \\
\textbf{$steps$} & 8 & 8 & 8 & 8 \\
\textbf{$stepsize$} & 2 & 20 & 2 & 2 \\
\hline
\end{tabular}
\end{threeparttable}
\end{center}
\caption{Parameters of FGSM and PGD.}
\label{tab:pgdsettings}
\end{table}

\subsection{Effects of BEFB on model training}\label{sec:EXTraining}
\begin{figure}[h!]
\centering
\subfigure[VGG16]{\includegraphics[width=0.4\textwidth]{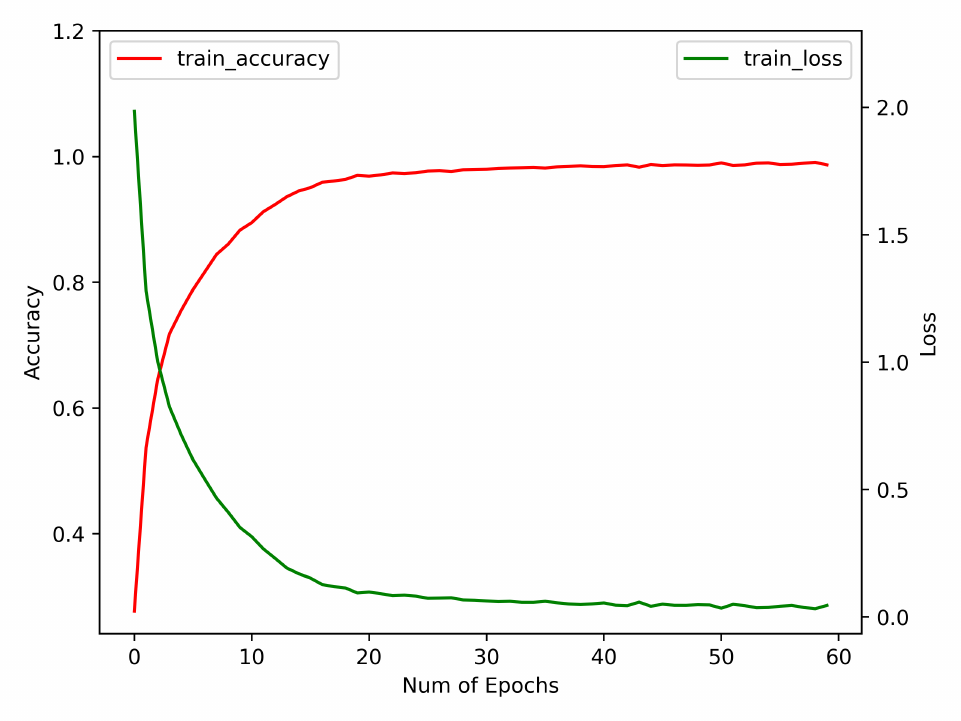}\label{fig:ORIBI}}
\hspace{0.05\textwidth}
\subfigure[VGG16-BEFB-multiple]{\includegraphics[width=0.4\textwidth]{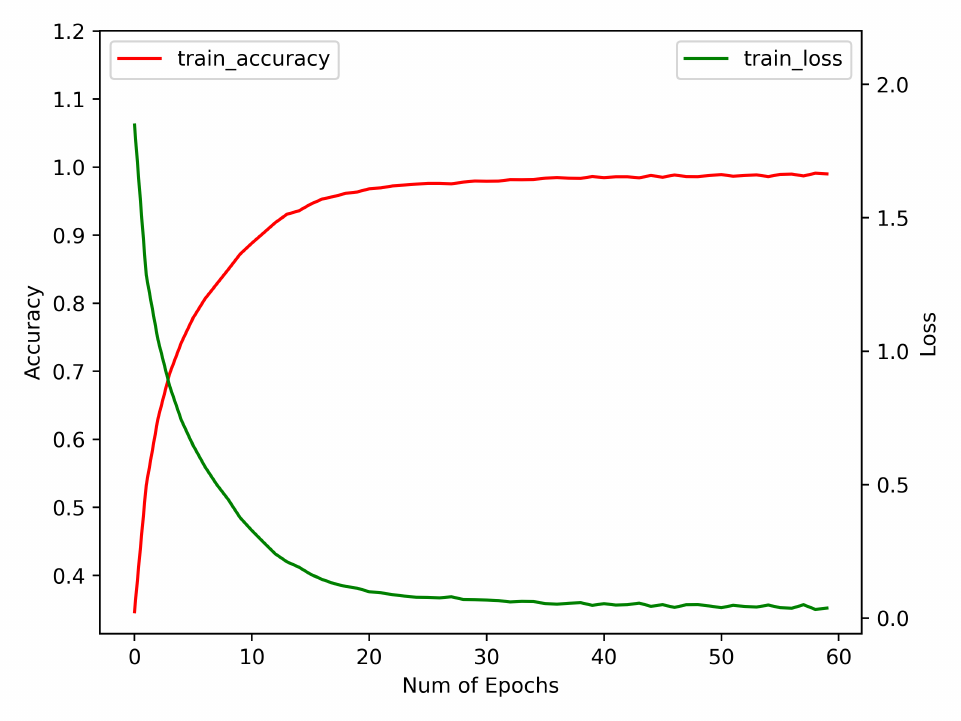}\label{fig:AEBI}}
\caption{\textbf{Comparison of training profiles between VGG16-BEFB-mutiple model and the original one on CIFAR-10 dataset}.}
\label{fig:cifar10trainingVGG}
\end{figure}

\begin{figure}[h!]
\centering
\subfigure[ResNet34]{\includegraphics[width=0.4\textwidth]{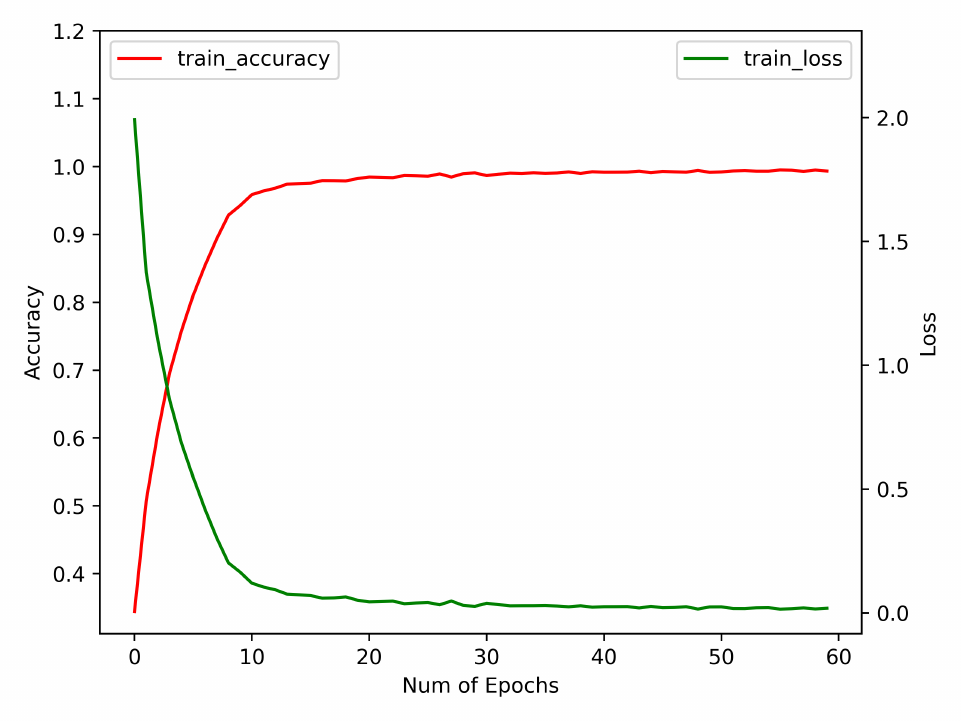}\label{fig:ORIBI}}
\hspace{0.05\textwidth}
\subfigure[ResNet34-BEFB-multiple]{\includegraphics[width=0.4\textwidth]{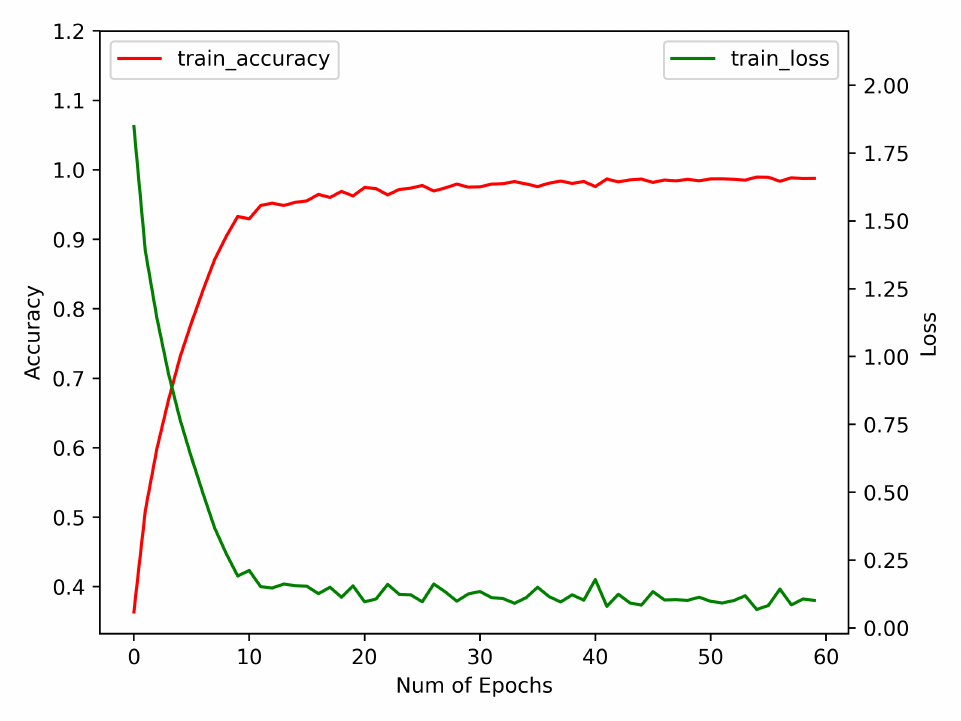}\label{fig:AEBI}}
\caption{\textbf{Comparison of training profiles between ResNet34-BEFB-mutiple model and the original one on CIFAR-10 dataset}.}
\label{fig:cifar10trainingRes}
\end{figure}

We examine the effects of BEFB on model training by comparing three metrics between the BEFB-multiple models and the original ones, i.e., training accuracy, test accuracy, and training time per epoch.
The loss function, optimizer, batch size, and number of epochs are set to be the same.
Table~\ref{tab:effectsBEFB} shows the comparison of training performance between BEFB-multiple models and the original ones.
From the table, it is clear to see the BEFB-multiple models are on a par with the original models on three metrics, which indicates BEFB is lightweight and has no side effects on model training.
Fig.~\ref{fig:cifar10trainingVGG} and Fig.~\ref{fig:cifar10trainingRes} depict the training profiles of VGG16-BEFB-multiple and ResNet34-BEFB-multiple on CIFAR-10 dataset, respectively. From the figures, we can see BEFB-multiple models demonstrate the similar training dynamics with the original ones.

\begin{table}[ht]
\hspace{-2.8cm}
\scriptsize
\begin{threeparttable}

\begin{tabular}{l|p{0.06\textwidth}p{0.06\textwidth}p{0.06\textwidth}|p{0.06\textwidth}p{0.06\textwidth}p{0.06\textwidth}|p{0.06\textwidth}p{0.06\textwidth}p{0.06\textwidth}|p{0.06\textwidth}p{0.06\textwidth}p{0.06\textwidth}}
\hline
\multirow{2}*{} & \multicolumn{3}{|c|}{\textbf{CIFAR-10}} & \multicolumn{3}{|c|}{\textbf{MNIST}} & \multicolumn{3}{|c}{\textbf{SVHN}} & \multicolumn{3}{|c}{\textbf{TinyIN}} \\
		\cline{2-13}
		~ & \textbf{Tr.Acc.} & \textbf{Te.Acc.} & \textbf{Ti.PE} & \textbf{Tr.Acc.} & \textbf{Te.Acc.} & \textbf{Ti.PE} & \textbf{Tr.Acc.} & \textbf{Te.Acc.} & \textbf{Ti.PE} & \textbf{Tr.Acc.} & \textbf{Te.Acc.} & \textbf{Ti.PE} \\
\hline
\textbf{VGG16} & 99.17\% & 83.70\% & 18s & 99.72\% & 99.40\% & 16s & 99.68\% & 94.36\% & 18s & 94.32\% & 33.84\% & 80s \\
\textbf{VGG16-BEFB-multiple} & 99.53\% & 82.95\% & 18s & 99.56\% & 99.46\% & 17s & 99.35\% & 94.64\% & 18s & 92.03\% & 30.01\% & 95s \\
\hline
\textbf{ResNet34} & 99.32\% & 79.19\% & 48s & 99.63\% & 99.40\% & 40s & 99.76\% & 93.22\% & 50s & 97.53\% & 30.80\% & 238s \\
\textbf{ResNet34-BEFB-multiple} & 99.48\% & 74.58\% & 48s & 99.38\% & 99.28\% & 42s & 99.78\% & 93.41\% & 52s & 97.98\% & 26.31\% & 250s \\
\hline
\end{tabular}
\begin{tablenotes}
\item \textbf{Tr.Acc.} stands for training accuracy. \textbf{Te.Acc.} stands for test accuracy. \textbf{Ti.PE} stands for training time per epoch.
\end{tablenotes}

\end{threeparttable}
\caption{Comparison of training performance between BEFB-multiple models and the original ones.}
\label{tab:effectsBEFB}
\end{table}

\subsection{Analysis of obfuscated gradient effect}
The activation function of the threshold layer is expressed by Eq.~(\ref{eq:featureoutput}), whose gradient is zero. In order to update the weights in BEFB, STE technique \cite{Bengio2013Arxiv, Yang2022ICML, Vanderschueren2023WACV} is adopted. But with respect to generating AEs, STE may bring obfuscated gradient effect \cite{Athalye2018ICML, Yue2023USENIX, Huang2022MSML}, which means
the yielded AEs are not powerful enough to deceive the models. Here, in addition to STE, zero gradient and gradient of a center-translated sigmoid activation function for threshold layer are also tested. Table~\ref{tab:obfuscategra} compares the classification accuracy of AEs generated by these three gradient computing strategies under FGSM attacks.
Eq.~(\ref{eq:sigmoid}) defines the center-translated sigmoid function. $x_{0}$ in general, is the threshold value for channels, illustrated in Eq.~(\ref{eq:featureoutput}).

\begin{table}[ht]
\scriptsize
\begin{center}
\begin{threeparttable}

\begin{tabular}{l|l|c|c|c|c}
\hline
\multicolumn{2}{l|}{} & \textbf{CIFAR-10} & \textbf{MNIST} & \textbf{SVHN} & \textbf{TinyIN} \\
\hline
\multirow{3}{*}{\textbf{VGG16-BEFB-multiple}} & \textbf{STE} & 64.24\% & 81.65\% & 71.02\% & 43.21\% \\
 & \textbf{Sigmoid} & 44.93\% &  68.25\% & 59.75\% & 33.06\% \\
 & \textbf{Zero Gradient} & 44.40\% & 67.42\% & 59.28\% & 31.97\% \\
\hline
\multirow{3}{*}{\textbf{ResNet34-BEFB-multiple}} & \textbf{STE} & 23.35\% & 17.55\% & 14.36\% & 35.87\% \\
 & \textbf{Sigmoid} & 14.87\% & 13.84\% & 12.71\% & 5.82\% \\
 & \textbf{Zero Gradient} & 14.88\% & 14.53\% & 12.17\% & 5.39\% \\
\hline
\end{tabular}
\end{threeparttable}
\end{center}
\caption{Comparison of classification accuracy of AEs generated by three gradient computing strategies under FGSM attacks.}
\label{tab:obfuscategra}
\end{table}

\begin{equation}
\label{eq:sigmoid}
f(x) = \frac{1}{1+e^{x-x_{0}}}
\end{equation}

From the table~\ref{tab:obfuscategra}, it is clear to see the classification accuracy under STE is much higher than zero gradient and gradient of center-translated sigmoid function. It indeed brings the obfuscated gradient effect using STE. In our experiments, we employ zero gradient of threshold layer to generate AEs.

\subsection{Performance comparison between BEFB integrated models and original models under attacks}
Table~\ref{tab:comparisonAttack} compares the classification accuracy of BEFB integrated models and the original models under FGSM, PGD, and C\&W attacks.

\begin{table}[ht]
\hspace{-2.8cm}
\scriptsize
\begin{threeparttable}

\begin{tabular}{l|p{0.06\textwidth}p{0.06\textwidth}p{0.06\textwidth}|p{0.06\textwidth}p{0.06\textwidth}p{0.06\textwidth}|p{0.06\textwidth}p{0.06\textwidth}p{0.06\textwidth}|p{0.06\textwidth}p{0.06\textwidth}p{0.06\textwidth}}
\hline
\multirow{2}*{} & \multicolumn{3}{|c|}{\textbf{CIFAR-10}} & \multicolumn{3}{|c|}{\textbf{MNIST}} & \multicolumn{3}{|c}{\textbf{SVHN}} & \multicolumn{3}{|c}{\textbf{TinyIN}} \\
		\cline{2-13}
		~ & \textbf{FGSM} & \textbf{PGD} & \textbf{C\&W} & \textbf{FGSM} & \textbf{PGD} & \textbf{C\&W} & \textbf{FGSM} & \textbf{PGD} & \textbf{C\&W} & \textbf{FGSM} & \textbf{PGD} & \textbf{C\&W} \\
\hline
\textbf{VGG16} & 28.59\% & 1.93\% & 0.06 & 56.54\% & 3.63\% & 6.78 & 57.77\% & 16.84\% & 0.08 & 27.39\% & 4.03\% & 0.05 \\
\textbf{VGG16-BEFB-single} & 42.77\% & 8.01\% & 0.05 & 59.05\% & 5.83\% & 6.53 & 58.15\% & 17.98\% & 0.08 & 31.15\% & 8.74\% & \textbf{0.07} \\
\textbf{VGG16-BEFB-multiple} & \textbf{45.60}\% & \textbf{8.05}\% & \textbf{0.07} & \textbf{67.42}\% & \textbf{15.53}\% & \textbf{7.03} & \textbf{59.28\%} & \textbf{19.48}\% & \textbf{0.10} & \textbf{31.93}\% & \textbf{9.18}\% & \textbf{0.07} \\
\hline
\textbf{ResNet34} & 7.82\% & 0.47\% & 0.05 & 12.66\% & 0.76\% & 1.22 & 24.14\% & 2.49\% & 0.02 & 5.80\% & 0.00\% & 0.10 \\
\textbf{ResNet34-BEFB-single} & 10.31\% & 0.78\% & 0.06 & 13.33\% & 2.33\% & \textbf{1.50} & 23.95\% & 2.41\% & 0.02 & \textbf{5.99}\% & 0.36\% & 0.11 \\
\textbf{ResNet34-BEFB-multiple} & \textbf{14.88}\% & \textbf{1.01}\% & \textbf{0.10} & \textbf{14.53}\% & \textbf{2.95}\% & 1.39 & \textbf{25.17}\% & \textbf{2.98}\% & \textbf{0.03} & 5.95\% & \textbf{0.95}\% & \textbf{0.14} \\
\hline
\end{tabular}
\end{threeparttable}
\caption{Comparison of classification accuracy between BEFB integrated models and original ones under attacks.}
\label{tab:comparisonAttack}
\end{table}

From the table, it is clear to see that BEFB-single and BEFB-multiple models are more robust than the original models under FGSM, PGD and C\&W attacks, and classification accuracy of BEFB-multiple models are slightly better than BEFB-single models. On CIFAR-10 dataset, VGG16-BEFB-multiple model can achieve 17\% and 6\% higher accuracy than the original model under FGSM and PGD attacks, respectively. On MNIST dataset, VGG16-BEFB-multiple model can achieve 11\% and 12\% higher accuracy than the original model under FGSM and PGD attacks, respectively.

\begin{figure}[h!]
    \centering
    \includegraphics[width=1\textwidth]{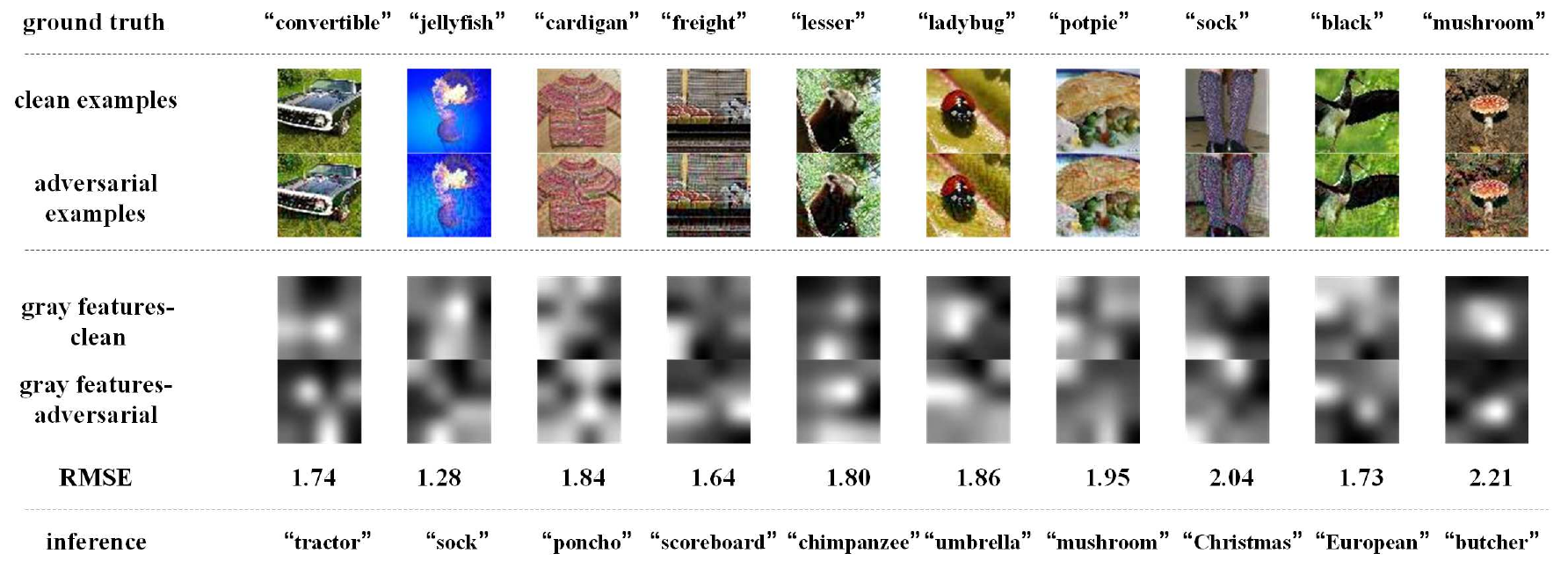}
    \caption{\textbf{Texture features extracted by the original VGG16 model and its predictions under FGSM of $\epsilon$=16 attack.}}
    \label{fig:featureFGSM16pureVGG}
\end{figure}

\begin{figure}[h!]
    \centering
    \includegraphics[width=1\textwidth]{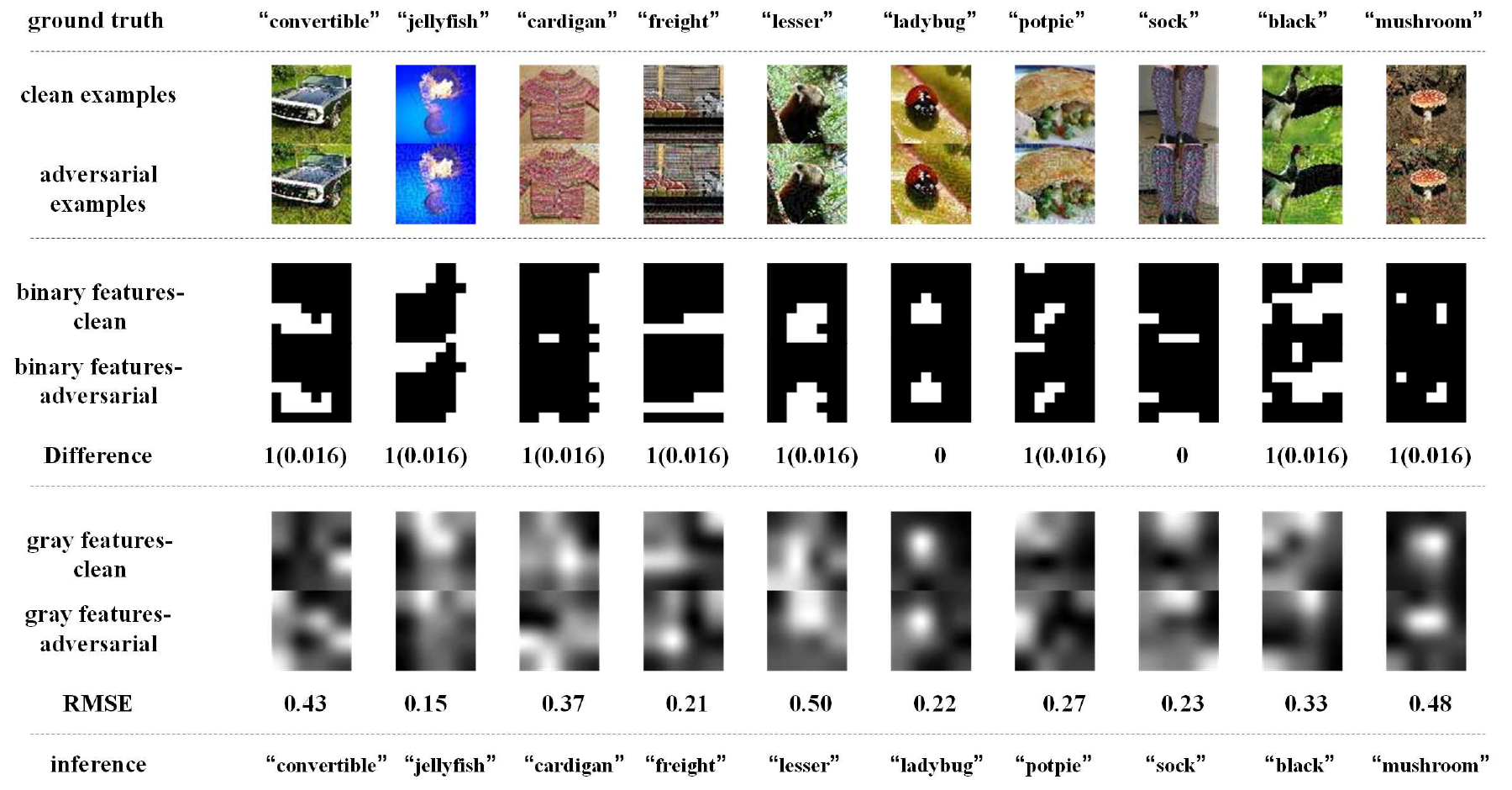}
    \caption{\textbf{Binary edge features and the texture features extracted by the VGG16-BEFB-multiple model and its predictions under FGSM of $\epsilon$=16 attack.}}
    \label{fig:featureFGSM16binaryVGG}
\end{figure}

\begin{figure}[h!]
    \centering
    \includegraphics[width=1\textwidth]{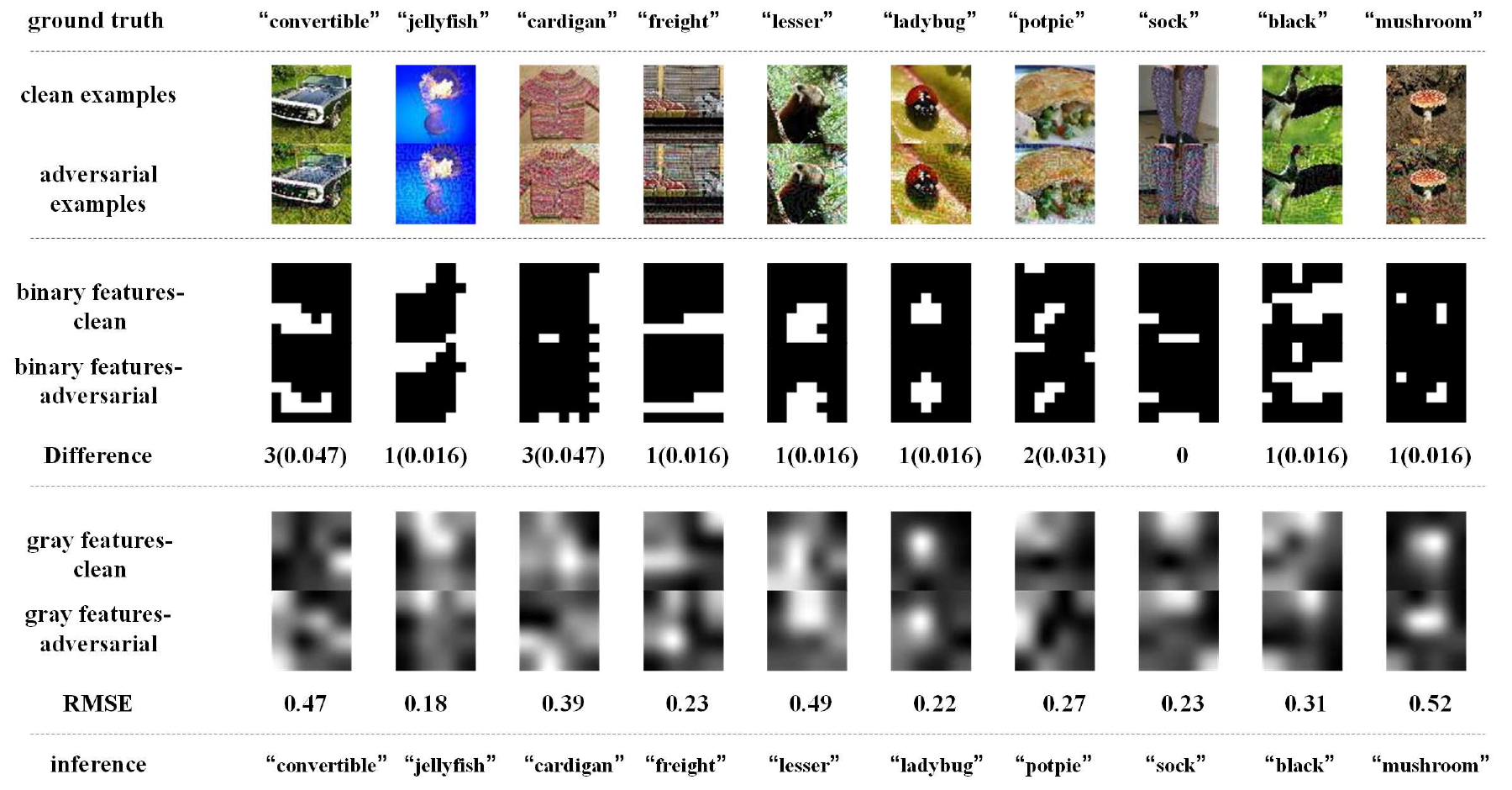}
    \caption{\textbf{Binary edge features and the texture features extracted by the VGG16-BEFB-multiple model and its predictions under FGSM of $\epsilon$=20 attack.}}
    \label{fig:featureFGSM20binaryVGG}
\end{figure}

\begin{figure}[h!]
    \centering
    \includegraphics[width=1\textwidth]{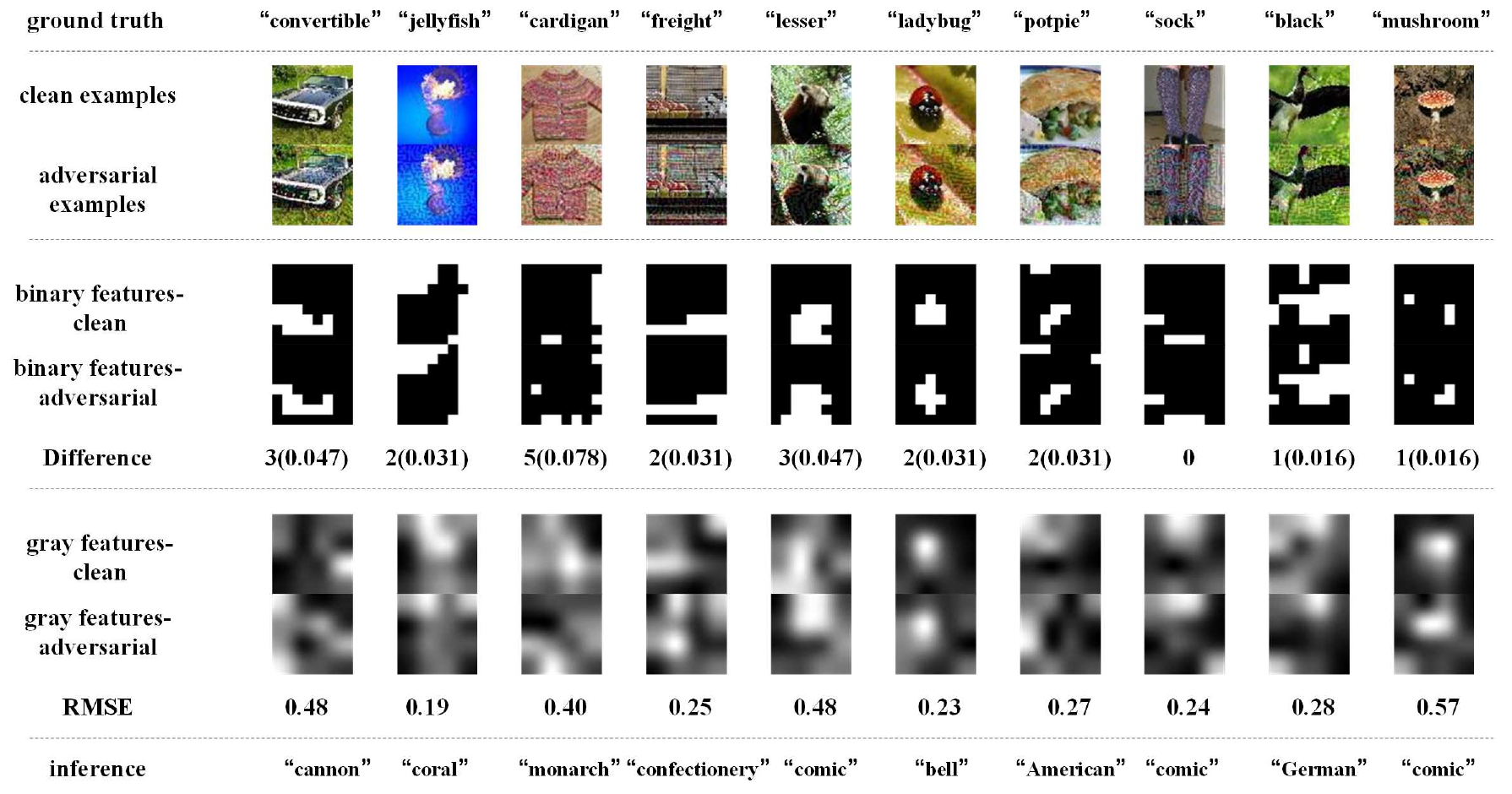}
    \caption{\textbf{Binary edge features and the texture features extracted by the VGG16-BEFB-multiple model and its predictions under FGSM of $\epsilon$=24 attack.}}
    \label{fig:featureFGSM24binaryVGG}
\end{figure}

Fig~\ref{fig:featureFGSM16pureVGG} and Fig~\ref{fig:featureFGSM16binaryVGG} examine the extracted features of both clean examples and the AEs by the original VGG16 model and VGG16-BEFB-multiple model on TinyIN dataset, respectively. On Fig.~\ref{fig:featureFGSM16pureVGG}, the certain images from TinyIN dataset under FGSM of $\epsilon$=16 attack are predicted wrong by the original VGG16 model, and it is clear to see there is significant difference between the extracted texture features of the clean examples and the AEs. On Fig.~\ref{fig:featureFGSM16binaryVGG}, the same images from TinyIN dataset under FGSM of $\epsilon$=16 attack are classified correctly by the VGG16-BEFF-multiple model, and the difference of the extracted texture features between the clean examples and the AEs is lower than that in the original model. More notably, the extracted binary edge features of clean examples and AEs are almost the same, e.g., in "ladybug" image and "sock" image, the number of different pixels are both zero. It indicates these extracted binary edge features play an important role in improving the classification accuracy on AEs. Furthermore, we add more perturbations to the images and observe the extracted features and prediction results of the VGG16-BEFB-multiple models. On Fig.~\ref{fig:featureFGSM20binaryVGG}, four more perturbations are added on the images under FGSM attack. The difference of texture features between the clean examples and AEs is slightly higher than that under $\epsilon$=16 attack, e.g., RMSE (root mean square error) of "jellyfish" image is 0.15 under $\epsilon$=16 and 0.18 under $\epsilon$=20, and RMSE of "mushroom" image is 0.48 under $\epsilon$=16 and 0.52 under $\epsilon$=20. The difference of binary edge features between the clean examples and AEs is nearly unchanged under $\epsilon$=16 attack and $\epsilon$=20 attack, e.g., the number of different pixels for "jellyfish" image is 1 under both $\epsilon$=16 attack and $\epsilon$=20 attack, and the number of different pixels for "mushroom" image is also 1 under both $\epsilon$=16 attack and $\epsilon$=20 attack. And the prediction results under FGSM of $\epsilon$=20 attack are correct. On Fig.~\ref{fig:featureFGSM24binaryVGG}, another four perturbations are added. And it can be seen that the difference of texture features under $\epsilon$=24 attack is slightly higher than that under $\epsilon$=20 attack, e.g., RMSE of "jellyfish" image is 0.18 under $\epsilon$=20 and 0.19 under $\epsilon$=24, and RMSE of "mushroom" image is 0.52 under $\epsilon$=20 attack and 0.57 under $\epsilon$=24 attack. The number of different pixels in binary edge features of clean examples and AEs keeps very close under $\epsilon$=20 attack and $\epsilon$=24 attack, e.g. the number for "mushroom" image keeps unchanged and the number for "jellyfish" image increases one.
Both Fig.~\ref{fig:featureFGSM20binaryVGG} and Fig.~\ref{fig:featureFGSM24binaryVGG} demonstrate the binary edge features are not susceptible to small perturbations, and to further improve the classification accuracy on AEs, exploring the novel combination forms for both texture features and binary edge features may be a key issue.

\begin{figure}[h!]
    \centering
    \includegraphics[width=1\textwidth]{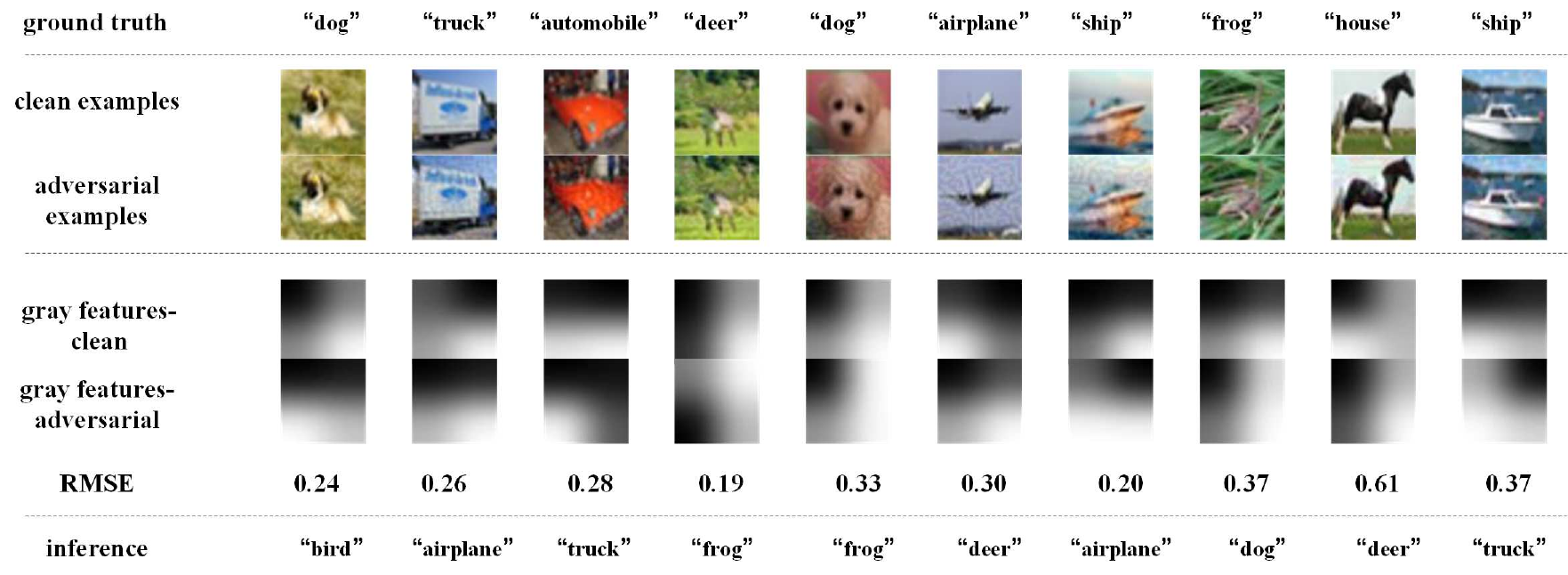}
    \caption{\textbf{Texture features extracted by the original ResNet34 model and its predictions under FGSM of $\epsilon$=8 attack.}}
    \label{fig:featureFGSM8pureRes}
\end{figure}

\begin{figure}[h!]
    \centering
    \includegraphics[width=1\textwidth]{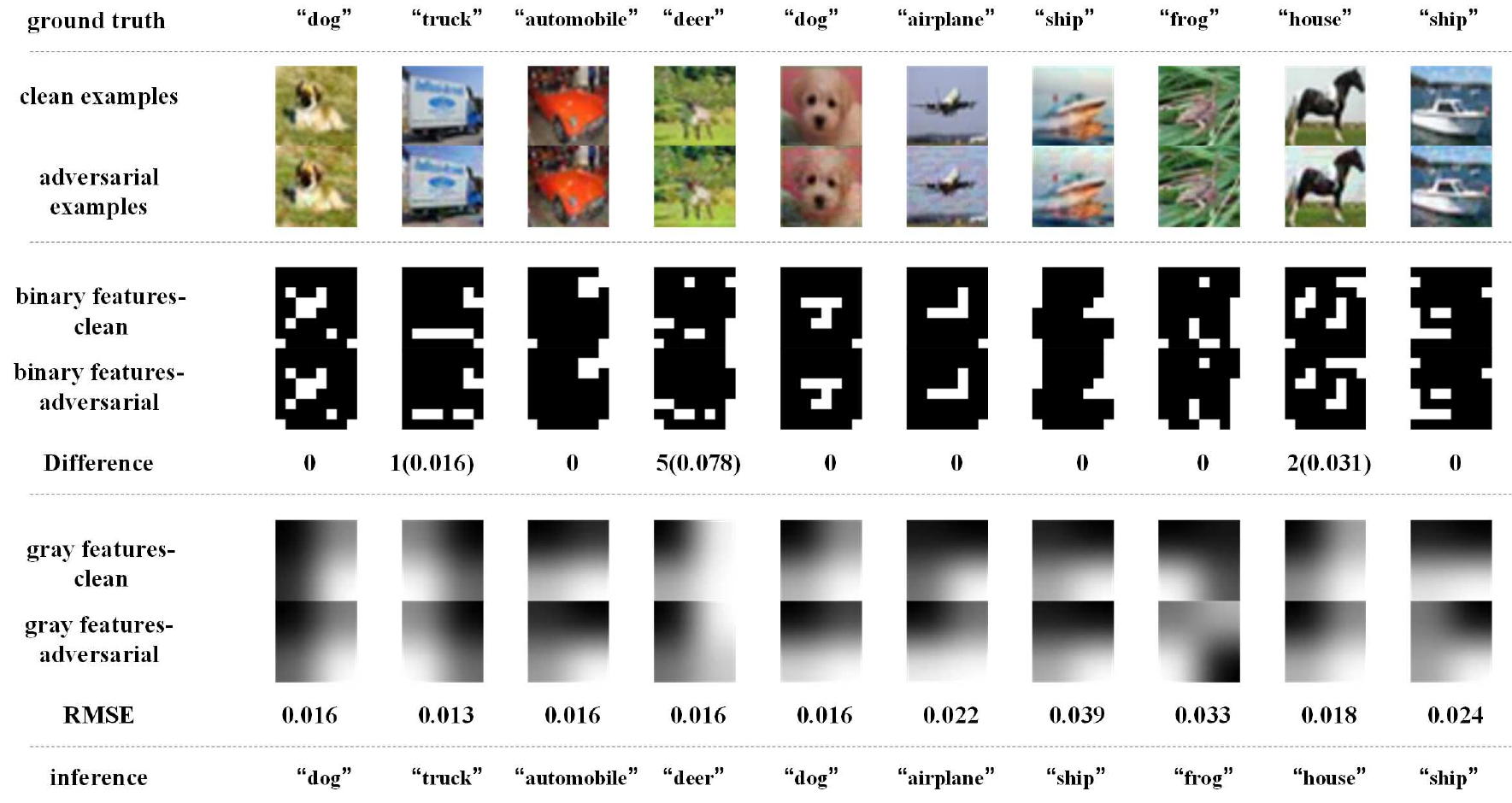}
    \caption{\textbf{Binary edge features and the texture features extracted by the ResNet34-BEFB-multiple model and its predictions under FGSM of $\epsilon$=8 attack.}}
    \label{fig:featureFGSM8binaryRes}
\end{figure}

\begin{figure}[h!]
    \centering
    \includegraphics[width=1\textwidth]{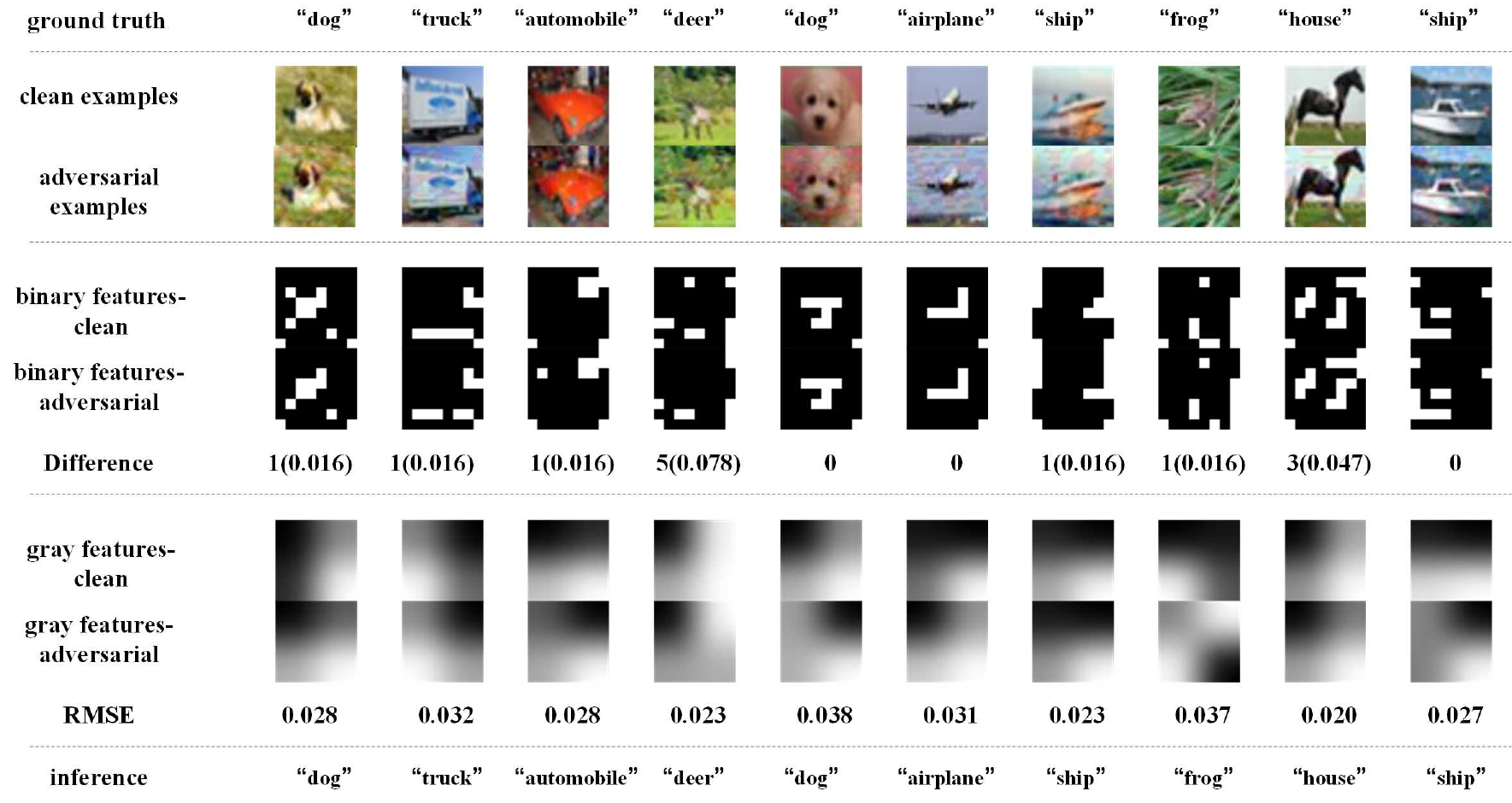}
    \caption{\textbf{Binary edge features and the texture features extracted by the ResNet34-BEFB-multiple model and its predictions under FGSM of $\epsilon$=12 attack.}}
    \label{fig:featureFGSM12binaryRes}
\end{figure}

\begin{figure}[h!]
    \centering
    \includegraphics[width=1\textwidth]{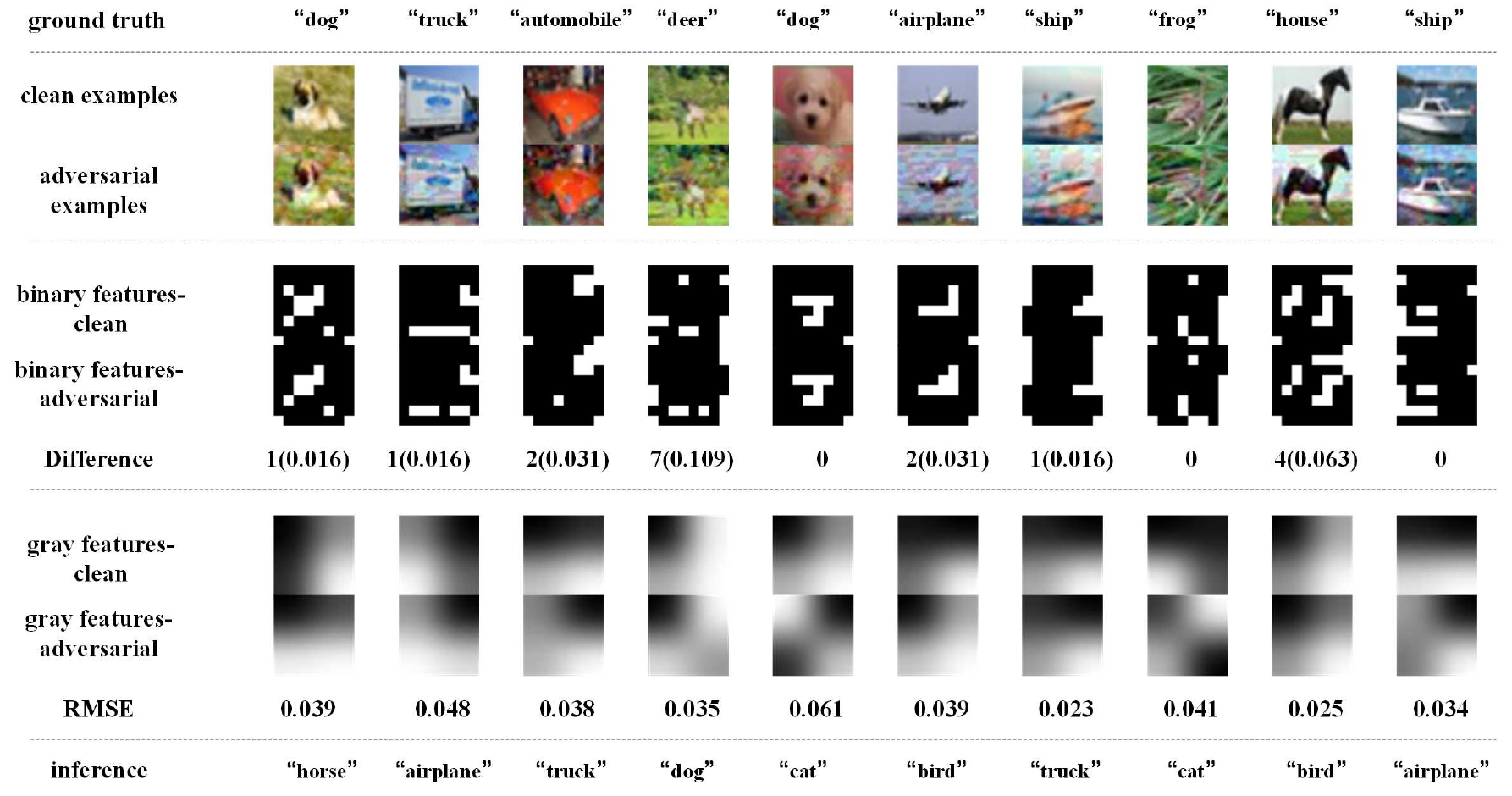}
    \caption{\textbf{Binary edge features and the texture features extracted by the ResNet34-BEFB-multiple model and its predictions under FGSM of $\epsilon$=16 attack.}}
    \label{fig:featureFGSM16binaryRes}
\end{figure}

Similarly, from Fig.~\ref{fig:featureFGSM8pureRes} to Fig.~\ref{fig:featureFGSM16binaryRes}, we examine the extracted features and prediction results for both clean examples and AEs by the original ResNet34 model and ResNet34-BEFB-multiple model on CIFAR-10 dataset. On Fig.~\ref{fig:featureFGSM8pureRes} and Fig.~\ref{fig:featureFGSM8binaryRes}, it can be seen that, the original ResNet34 model makes the wrong predictions for the certain images from CIFAR-10 dataset under FGSM of $\epsilon$=8 attack, while ResNet34-BEFB-multiple model gets them correct because of the extracted binary edge features almost the same between the clean examples and the AEs. For example, on "dog", "airplane", "automobile", "ship" and "frog" images, the number of different pixels is zero.
On Fig.~\ref{fig:featureFGSM12binaryRes} and Fig.~\ref{fig:featureFGSM16binaryRes}, four more and eight more perturbations are added, respectively. It can be seen that the prediction results are correct under FGSM of $\epsilon$=12 attack and wrong under FGSM of $\epsilon$=16 attack.
When focusing on the binary edge features under these two attacks, it is found the changes on binary edge features in most of cases are zeros, e.g., in "dog", "truck", and "ship" images.

\begin{figure}[h!]
\centering
\subfigure[Gaussian noise with zero mean and 0.08 standard deviation on CIFAR-10 dataset]{\includegraphics[width=0.4\textwidth]{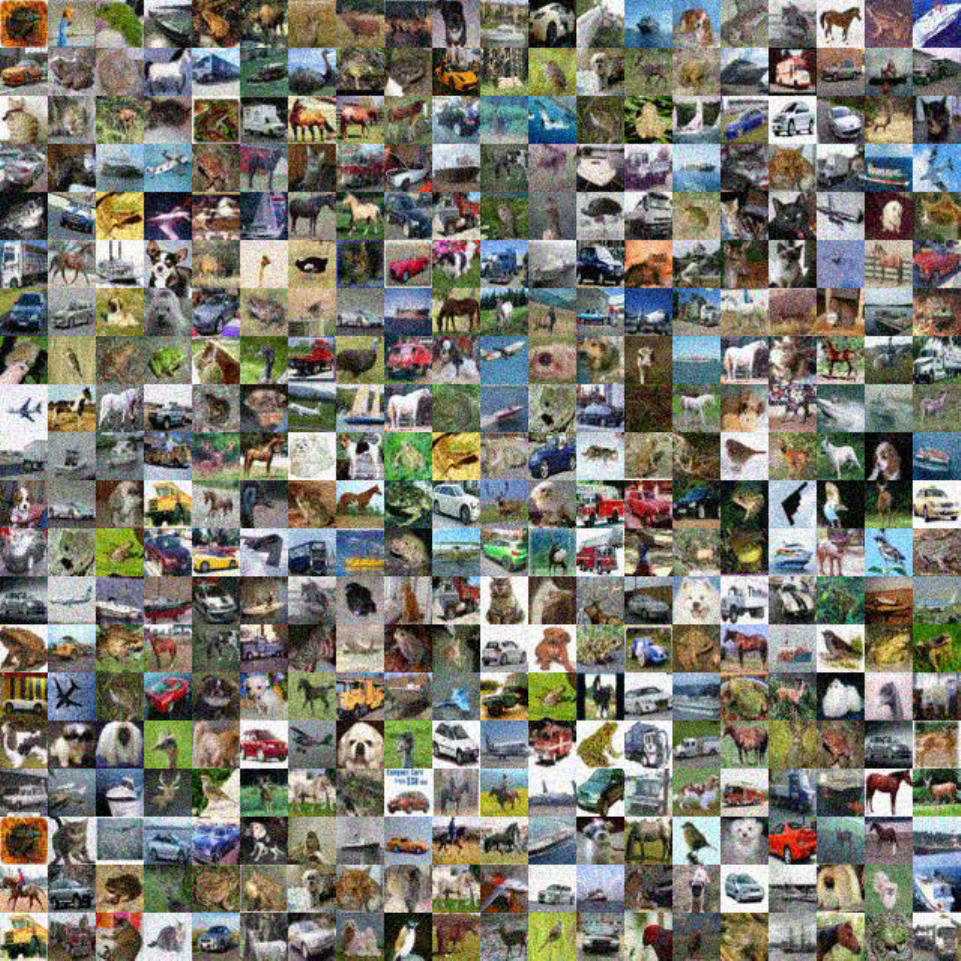}\label{fig:gausssiannoiseCIFAR}}
\hspace{0.05\textwidth}
\subfigure[Gaussian noise with zero mean and 0.35 standard deviation on MNIST dataset]{\includegraphics[width=0.4\textwidth]{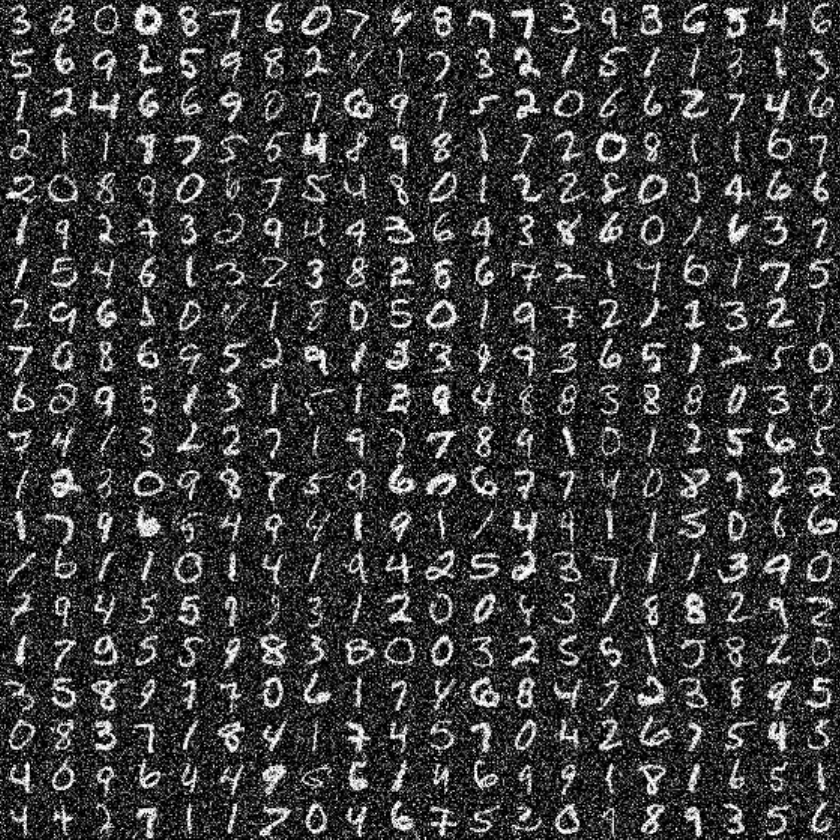}\label{fig:gausssiannoiseMNIST}}
\caption{\textbf{Gaussian noise perturbed images}.}
\label{fig:gausssiannoiseCIFARMNIST}
\end{figure}

We also compare the classification performance of both original models and BEFB integrated models on the images perturbed by gaussian noise.
Fig.~\ref{fig:gausssiannoiseCIFARMNIST} illustrates the gaussian noise perturbed images of CIFAR-10 dataset and MNIST dataset.
In table~\ref{tab:gaussiannoisecom}, it is clear to see the classification accuracy of BEFB integrated models are better than the original models on both datasets, e.g. the ResNet34-BEFB-multiple model can achieve 15\% and 10\% higher accuracy than the original model on CIFAR-10 dataset and MNIST dataset, respectively.

%
%

\begin{table}[ht]
\scriptsize
\begin{center}
\begin{threeparttable}

\begin{tabular}{l|c|c}
\hline
~ & \textbf{CIFAR-10} & \textbf{MNIST} \\
\hline
\textbf{VGG16} & 61.35\% & 85.17\% \\
\textbf{VGG16-BEFB-multiple} & \textbf{64.17}\% & \textbf{88.75}\% \\
\hline
\textbf{ResNet34} & 75.63\% & 76.01\%\\
\textbf{ResNet34-BEFB-multiple} & \textbf{91.62}\% & \textbf{86.11}\%\\
\hline
\end{tabular}
\end{threeparttable}
\caption{Comparison of classification accuracy between the original models and BEFB integrated models on gaussian noise perturbed CIFAR-10 and MNIST datasets.}
\label{tab:gaussiannoisecom}
\end{center}
\end{table}

\subsection{Combining BEFB integrated models with AT and PCL}
\begin{figure}[h!]
    \centering
    \includegraphics[width=0.5\textwidth]{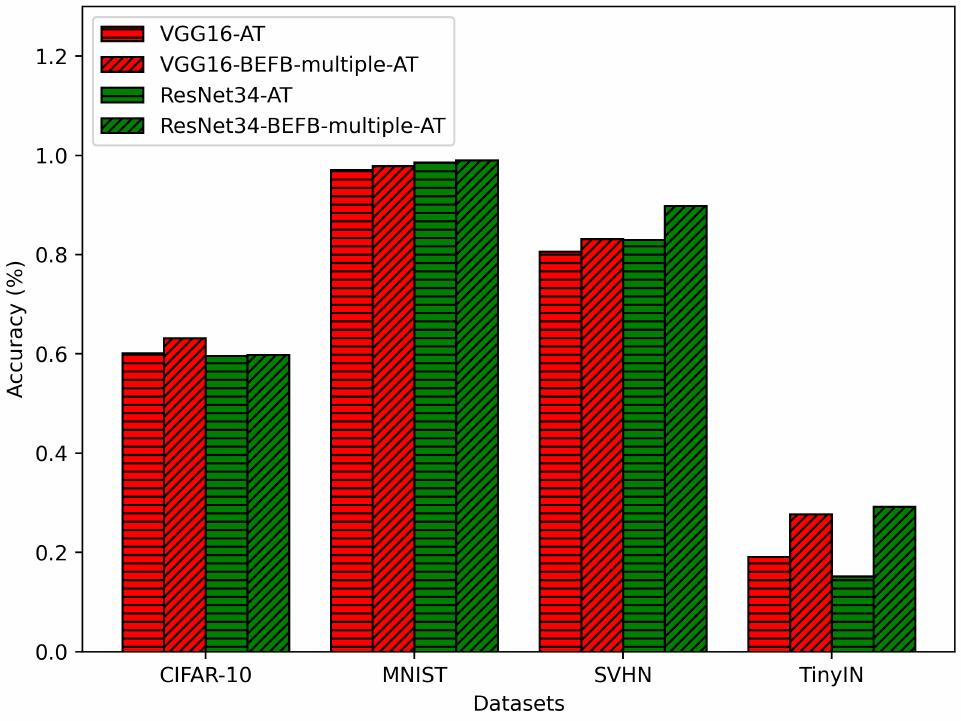}
    \caption{\textbf{Comparing AT enhanced BEFB-multiple models with original models under PGD attack.}}
    \label{fig:ATPGD}
\end{figure}

\begin{figure}[h!]
\centering
\subfigure[under FGSM attack]{\includegraphics[width=0.45\textwidth]{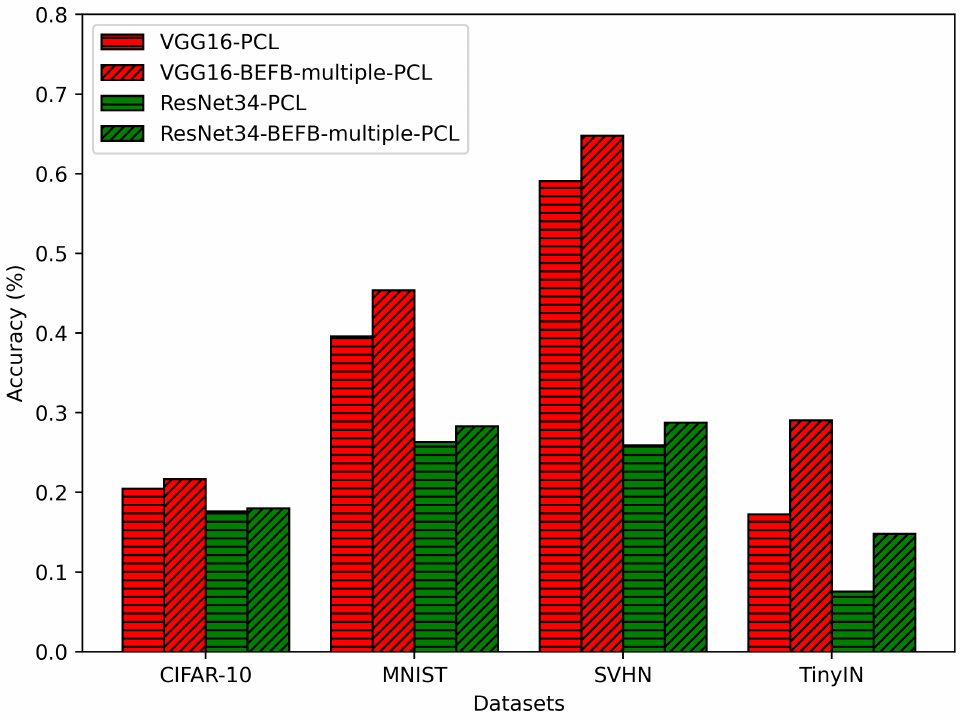}\label{fig:PCLFGSM}}
\hspace{0.05\textwidth}
\subfigure[under PGD attack]{\includegraphics[width=0.45\textwidth]{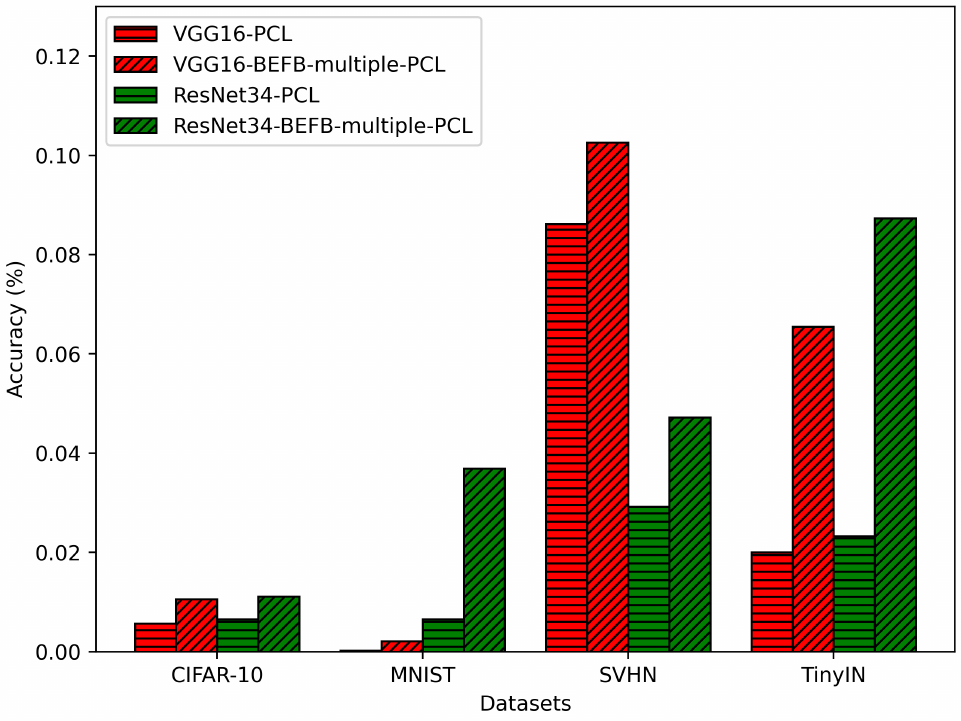}\label{fig:PCLPGD}}
\caption{\textbf{Comparing PCL enhanced BEFB-multiple models with original models under FGSM and PGD attacks}.}
\label{fig:PCLFGSMPGD}
\end{figure}

We combine BEFB-multiple models with two popular robustness enhancing techniques-\--AT and PCL, and compare them to the original models with AT and PCL. We denote VGG16-BEFB-multiple models with AT as VGG16-BEFB-multiple-AT, and VGG16-BEFB-multiple models with PCL as VGG16-BEFB-multiple-PCL. The same notations for ResNet34-BEFB-multiple models and the original models.
Fig~\ref{fig:ATPGD} compares the classification accuracy of BEFB-multiple-AT models with AT enhanced original models under PGD attack. Fig.~\ref{fig:PCLFGSMPGD} compares the classification accuracy of BEFB-multiple-PCL models with PCL enhanced original models under FGSM and PGD attacks.

%
From the figures, we can see AT and PCL enhanced BEFB-multiple models have better classification accuracy than AT and PCL enhanced original models.

\subsection{Ablation study}
\begin{figure}[h!]
\centering
\subfigure[Edge images for clean examples]{\includegraphics[width=0.3\textwidth]{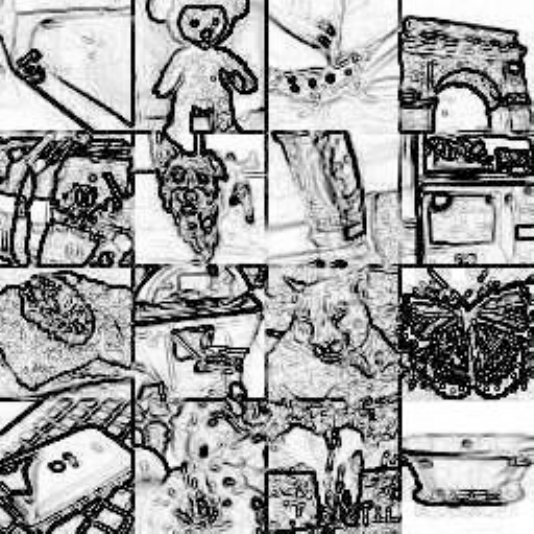}\label{fig:ORIEDGE}}
\hspace{0.05\textwidth}
\subfigure[Edge images for adversarial examples]{\includegraphics[width=0.3\textwidth]{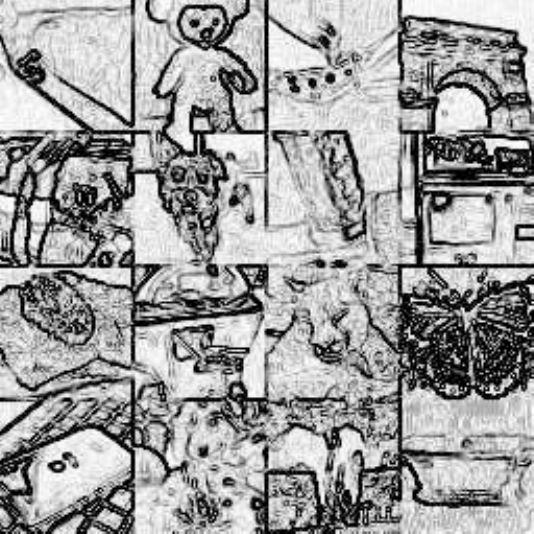}\label{fig:AEEDGE}}
\caption{\textbf{Edge images of clean examples and adversarial examples}.}
\label{fig:edgeimages}
\end{figure}

BEFB integrated models have two main components, i.e., Sobel layers and threshold layer. We remove them respectively in BEFB-multiple models to observe the accuracy change. We denote threshold layer removed models as BEFB-multiple-tlre, and Sobel layers removed models as BEFB-multiple-slre.
Table~\ref{tab:ThresholdSobellayerremoved} shows the comparison of classification accuracy between the original BEFB-multiple models and BEFB-multiple models with threshold layer and Sobel layers removed, respectively. From the table, it is clear to see when removing threshold layer and Sobel layers, respectively, the robustness of BEFB-multiple-tlre and BEFB-multiple-slre models is weakened. This can be also illustrated by Fig.~\ref{fig:edgeimages}. Fig.~\ref{fig:AEEDGE} shows, without thresholding, the small noise in AEs is amplified after performing edge detection.

\begin{table}[ht]
\scriptsize
\begin{center}
\begin{threeparttable}

\begin{tabular}{l|c|c|c|c}
\hline
\multirow{2}*{} & \multicolumn{2}{|c|}{\textbf{CIFAR-10}} & \multicolumn{2}{|c}{\textbf{SVHN}} \\
		\cline{2-5}
		~ & \textbf{FGSM} & \textbf{PGD} & \textbf{FGSM} & \textbf{PGD}\\
\hline
\textbf{VGG16-BEFB-multiple-tlre} & 40.55\% & 6.73\% & 48.70\% & 10.46\% \\
\textbf{VGG16-BEFB-multiple-slre} & 33.73\% & 2.90\% & 55.22\% & 15.89\% \\
\textbf{VGG16-BEFB-multiple} & \textbf{45.60}\% & \textbf{8.05}\% & \textbf{59.28}\% & \textbf{19.48}\% \\
\hline
\textbf{ResNet34-BEFB-multiple-tlre} & 13.58\% & 0.09\% & 24.22\% & 2.44\% \\
\textbf{ResNet34-BEFB-multiple-slre} & 9.11\% & 0.86\% & 16.42\% & 2.18\% \\
\textbf{ResNet34-BEFB-multiple} & \textbf{14.88}\% & \textbf{1.01}\% & \textbf{25.17}\% & \textbf{2.98}\% \\
\hline
\end{tabular}
\end{threeparttable}
\caption{Comparison of removing threshold layer and Sobel layers respectively.}
\label{tab:ThresholdSobellayerremoved}
\end{center}
\end{table}

\section{Discussions}\label{sec:Dis}
The experimental results from Section~\ref{sec:Expe} shows BEFB has no side effects on model training, and BEFB integrated models are more robust than original models under attacks. And when combining BEFB integrated models with AT and PCL, it can achieve better classification accuracy than original models. It also can be seen that, using BEFB to enhance robustness is effective but not that significant. It may be because under BEFB, the binary edge features are combined with texture features by concatenation. We believe it is worthwhile to explore other combination forms of binary edge features and texture features which can potentially improve the robustness of DCNNs notably.

\section{Conclusions}\label{sec:Con}
Enhancing the robustness of DCNNs is of great significance for the safety-critical applications in the real world. Inspired by the principal way that human eyes recognize objects, in this paper, we design four edge detectors and propose a binary edge feature branch (BEFB for short), which can be easily integrated into any popular backbone. Experiments on multiple datasets show BEFB has no side effects on model training, and BEFB integrated models are more robust than the original models. The work in this paper for the first time shows it is feasible to combine shape-like features and texture features to make DCNNs more robust. In future's work, we endeavor to explore other effective and efficient combination forms of binary edge features and texture features, and design an optimization framework for the parameter searching to yield models with good performance under attacks.

\section*{Acknowledgment}
This work was supported by the National Natural Science Foundation of China (No.51977193).





\bibliographystyle{elsarticle-num}
\bibliography{egbib}

%





\end{document}